\documentclass[pmlr,twocolumn,10pt]{jmlr} % W&CP article

\usepackage{amsmath,amsfonts,bm}
\usepackage[normalem]{ulem}
\usepackage[dvipsnames]{xcolor}

\newcommand{\stkout}[1]{\ifmmode\text{\sout{\ensuremath{#1}}}\else\sout{#1}\fi}

\definecolor{mygray}{gray}{0.4}

\newcommand{\plusminus}[1] { \scriptsize{$\pm$ #1}}

\newcommand{\Denv} {\mathcal{D}_{\text{env}}}

\newcommand{\D} {\mathcal{D}}

\theoremheaderfont{\normalfont\bfseries}
\newtheorem{mytheorem}{Theorem}
\newtheorem{mylemma}{Lemma}
\theoremheaderfont{\normalfont\bfseries}
\newtheorem{assumption}{Assumption}
\newtheorem*{theorem*}{Theorem}

\def\eqref#1{eqn~\ref{#1}}
\def\1{\bm{1}}

\DeclareMathAlphabet{\mathsfit}{\encodingdefault}{\sfdefault}{m}{sl}
\SetMathAlphabet{\mathsfit}{bold}{\encodingdefault}{\sfdefault}{bx}{n}

\def\gA{{\mathcal{A}}}

\def\gD{{\mathcal{D}}}

\def\gF{{\mathcal{F}}}

\def\gM{{\mathcal{M}}}
\def\gN{{\mathcal{N}}}

\newcommand{\R}{\mathbb{R}}

\usepackage{multirow}
\usepackage{wrapfig} % wraptable
\usepackage{adjustbox}

\usepackage[utf8]{inputenc} % allow utf-8 input
\usepackage[T1]{fontenc} % use 8-bit T1 fonts 
\usepackage{hyperref} % hyperlinks 
\usepackage{url} % simple URL typesetting \usepackage{booktabs} % professional-quality tables 
\usepackage{amsfonts} % blackboard math symbols 
\usepackage{nicefrac} % compact symbols for 1/2, etc.
\usepackage{microtype} % microtypography 
\usepackage{xcolor} % colors 
\usepackage{amsmath,amssymb} 
\usepackage{algorithm} 
\usepackage{algpseudocode} 
\usepackage{float}
\usepackage{booktabs,graphicx,caption}
\usepackage{makecell} 
\usepackage{subcaption} % The lineno package is required for denoting line % numbers for paper review. 
\usepackage[switch]{lineno}

\newcommand{\equal}[1]{{\hypersetup{linkcolor=black}\thanks{#1}}}

\jmlrvolume{297} 
\jmlryear{2025}
\jmlrworkshop{Machine Learning for Health (ML4H) 2025} % W&CP title

\title[Guardian-regularized Safe Offline Reinforcement Learning for MCS Weaning]{Guardian-regularized Safe Offline Reinforcement Learning for Smart Weaning of Mechanical Circulatory Devices}

\author{%
\Name{Aysin Tumay}\equal{These authors contributed equally} \Email{atmay@ucsd.edu}\\
\addr University of California, San Diego
\AND
\Name{Sophia Sun}\footnotemark[1] \Email{shs066@ucsd.edu}\\
\addr University of California, San Diego
\AND
\Name{Sonia Fereidooni} \Email{sfereidooni@ucsd.edu}\\
\addr University of California, San Diego
\AND
\Name{Aaron Dumas} \Email{adumas@caltech.edu}\\
\addr California Institute of Technology
\AND
\Name{Elise Jortberg} \Email{ jortberg.e@gmail.com}\\
\addr Abiomed
\AND
\Name{Rose Yu} \Email{roseyu@ucsd.edu}\\
\addr University of California, San Diego
}

\begin{document}

\maketitle

\begin{abstract}
We study the sequential decision-making problem for automated weaning of mechanical circulatory support (MCS) devices in cardiogenic shock patients. MCS devices are percutaneous micro-axial flow pumps that provide left ventricular unloading and forward blood flow, but current weaning strategies vary significantly across care teams and lack data-driven approaches.  Offline reinforcement learning (RL) has proven to be successful in sequential decision-making tasks, but our setting presents challenges for training and evaluating traditional offline RL methods:
prohibition of online patient interaction, highly uncertain circulatory dynamics due to concurrent treatments, and limited data availability. 
We developed an end-to-end machine learning framework with two key contributions 
(1) \textbf{C}linically-aware \textbf{O}OD-\textbf{r}egularized \textbf{M}odel-based \textbf{P}olicy \textbf{O}ptimization (\textsf{CORMPO}) a density-regularized offline RL algorithm for out-of-distribution suppression
that also incorporates clinically-informed reward shaping and (2) a Transformer-based probabilistic digital twin that models MCS circulatory dynamics for policy evaluation with rich physiological and clinical metrics.
%(1) a Transformer-based probabilistic digital twin of MCS circulatory dynamics for policy evaluation, and (2) a density-regularized offline RL algorithm that incorporates clinically-informed reward shaping and uses density estimation for safe out-of-distribution regularization. 
We prove that \textsf{CORMPO} achieves theoretical performance guarantees under mild assumptions. \textsf{CORMPO} attains a higher reward than the offline RL baselines by 28\% and higher scores in clinical metrics by 82.6\% on real and synthetic datasets. 
%We provide theoretical performance guarantees under mild assumptions and demonstrate that CORMPO outperforms established offline RL baselines across clinically-relevant metrics. 
Our approach offers a principled framework for safe offline policy learning in high-stakes medical applications where domain expertise and safety constraints are essential. 
%We provide our code repository: \href{https://github.com/aysintumay/CORMPO}{https://github.com/aysintumay/CORMPO}.
\end{abstract}

\paragraph{Data and Code Availability.}  The code repository\footnote{available at \href{https://github.com/Rose-STL-Lab/CORMPO}{https://github.com/Rose-STL-Lab/CORMPO}} includes the full implementation of our method and the synthetic dataset. The real-world dataset contains human-subject information and cannot be publicly released.

\paragraph{Institutional Review Board (IRB).} This study used real patient data under IRB approval.

\section{Introduction}
%\soni{refresher - https://github.com/Rose-STL-Lab/Lab-Wiki/wiki/Paper-submission-checklist}

%Offline Reinforcement learning (RL) 

%However,  real-world RL applications are challenged by the lack of online interaction data, forcing reliance on offline datasets. This setting introduces two critical issues: distribution shift between historical data and the real environment, and uncertainty in transition rollouts.
%Medical decision-making supported with time series forecasting has been extensively studied to help physicians uncover intrinsic patterns in clinical data and develop accurate prognoses.   
%This task is well-formulated with time series forecasting and offline reinforcement learning (RL), both emerging as a powerful recipe for sequential decision-making, where traditional methods may fall short. 
%Learning sequential decisions with offline reinforcement learning (RL) from retrospective data offers a richer guidance to clinicians than rule-based guidelines. To answer ``what-if'' questions on patient data and simulate an RL environment for evaluation, we identify a critical need to develop probabilistic simulators.
%However, noise in clinical measurements, imbalanced data sampling, and stochasticity of the real-life decisions make this task challenging for offline RL policies and simulators.

%what is the task
%This paper focuses on the problem of data-driven automatic weaning of mechanical circulatory support (MCS) devices. %, specifically the Impella. The Impella device 
MCS devices assist the heart by pumping oxygen-rich blood from the left ventricle into the ascending aorta, supporting patients with compromised cardiac function.  Weaning from MCS is a series of flow controls over a period of time in which the clinician aims to reduce flow support while maintaining stable hemodynamics, prior to explanting the MCS device \citep{Atti2022ComprehensiveMCS}. 
%Clinicians adjust the pump flow level (P-Level) based on heuristic evaluations of patient condition. 
Reducing the pump flow level (P-Level) is entirely at the discretion of the clinician: the manufacturer's instructions for use suggest reducing by levels of 2, and evaluating at each reduction for evidence of deterioration. However, constant monitoring of the patient state is heuristics-based without any empirical or theoretical basis.
%This underscores the need for a decision-support system that draw on both historical patient states and probabilistic forecasts of future hemodynamic stability.
%-- an ``integrated assessment of clinical exam, metrics of end-organ perfusion, imaging, and invasive hemodynamic data’’ \citep{weaning}. %The high stakes associated with these decisions underscore the need for a data-driven policy that can safely guide the setting of the MCS flow level during the weaning process. %By reinforcing the safe regions of the states of the simulated environment and the actions at the p level, our approach aims to provide safer and more reliable recommendations for the management of these life support devices.

%why use RL and the limitations
%There has been a growing interest in using deep reinforcement learning (RL) for sequential decision making in medical treatment, 
Deep reinforcement learning (RL) has shown great promise in automating sequential decision making in medical treatments, with works exploring clinical conditions such as sepsis \citep{raghu2017continuous, komorowski2018artificial} and cancer \citep{tseng2017deep, eckardt2021reinforcement}.
%and control of surgical robots \citep{kweon2021deep, yang2022guidewire}. 
With RL's ability to learn sequential decisions from real-world datasets, a data-driven policy can reduce clinician decision fatigue and offer richer guidance compared to rule-based guidelines.  
%However, in the medical domain, offline RL introduces challenges in policy training and evaluation. Firstly, the learned policy suffers from the stochasticity of the clinician's decisions and small data size. Secondly, since online interaction with patients on MCS is infeasible, evaluation solely relies on simulators.
%that are inherently stochastic due to the noisy training data.
%However, in the medical domain, offline RL introduces two challenges: in training, the stochasticity of clinician decisions and limited data hinder learning; in evaluation, because online interaction with patients is infeasible, assessment must rely solely on simulators.

Our application acknowledges three challenges. First, medical treatments cannot be learned via online interaction or exploration on patients, nor can the learned policy be directly evaluated in the real world. 
Second, the dynamics of weaning MCS devices are highly uncertain and require clinical discretion: patients on MCS are likely also receiving other sources of treatment such as surgery and medications, which the devices's learning algorithm does not have access to. 
%Third, our dataset consists of 379 anonymized treatment trajectories of varying lengths, substantially smaller than the datasets used in common offline RL applications. 
Due to these challenges, state-of-the-art safe offline reinforcement learning algorithms such as uncertainty penalization \citep{yu2020mopo} or value regularization \citep{svr} becomes over-conservative and unstable in our setting.
%Previous work in offline RL offers uncertainty penalization in model-based rollouts \cite{yu2020mopo} and OOD-suppression in policy optimization \cite{svr} alongside employing a Double Deep Q-network for sepsis treatment. 

We propose an end-to-end  pipeline for learning clinically informed MCS weaning strategies. Our method, \textbf{C}linically-aware \textbf{O}OD-\textbf{r}egularized \textbf{M}odel-based \textbf{P}olicy \textbf{O}ptimization (\textsf{CORMPO}) tackles the challenges by leveraging a probabilistic digital twin for evaluation, incorporating domain-specific metrics to encourage medically salient behaviors, and developing a density-regularized offline RL algorithm to ensure policy safety while optimizing performance. We show both theoretical and empirical support for our method's strong performance. 
In summary, our contributions are: 
\begin{enumerate} 
 %that provides a continuous state-space representation for offline RL, quantifying the uncertainty inherent in its predictions. 

\item  We present a Markov Decision Process (MDP) formulation for learning MCS weaning with offline RL models. To evaluate the models, we develop domain-specific clinically-aware metrics and a transformer-based probabilistic digital twin that models MCS circulatory dynamics.
\item  Our offline RL algorithm 
%\textbf{C}linically-aware \textbf{O}OD-regularized \textbf{M}odel-based \textbf{P}olicy \textbf{O}ptimization 
\textsf{CORMPO} utilizes reward shaping and a novel density-based OOD (out-of-distribution) penalization method. The algorithm achieves performance guarantees under mild assumptions, and outperforms offline RL baselines by 28\% in physiological reward and by 82.6\% in clinical metrics on real and synthetic datasets. 
\end{enumerate}

While \textsf{CORMPO} is developed for the specific application of MCS weaning, our methodological contributions, including the density-based OOD safeguarding and clinically-informed reward design, are broadly applicable to many medical decision-making applications that face similar challenges as ours.

\vspace{-0.5em}
 \section{Related Work}
 %Our work builds on advances in safety-aware and uncertainty-aware offline reinforcement learning (RL) as well as medical decision making. 
%\subsection{Safety-aware offline RL}

\paragraph{Safety-aware Offline RL.} Early safety-aware offline RL methods \citep{liu2021safetyeditor} use a dual-policy framework to transform potentially unsafe actions, while \cite{yang2022regularizing} addressed distribution shifts by regularizing the stationary state–action distribution of the current policy to match that of the offline dataset.  \cite{ran2023prdc} proposed Policy Regularization with Dataset Constraint which explores near-dataset state–action pairs using a nearest-neighbor search. The single-neighbor selection may induce bias, a limitation we mitigate by predicting continuous probabilities for each state–action pair. \cite{svr} offers a simple and effective means of directly regularizing Q-value estimates by penalizing values outside the support of the behavior policy, and \cite{spot} pursues the same goal using density estimates of state–action pairs. \cite{cped} further advances this idea by employing a FlowGAN-based model for density estimation. However, these safety-focused methods struggle to generalize to real-world datasets, where small data, high dimensionality, and environment stochasticity undermines their effectiveness and stability. %reliance on density models and behavior policies is undermined by sparse coverage.

\paragraph{Learning for safe medical decision making.}  \cite{prasad2018weaning} pioneered the application of RL for weaning mechanical ventilation, yet their fitted Q-iteration approach struggled with suboptimal clinical data. \cite{tang2018srlrnn} leveraged recurrent neural networks to capture temporal dependencies for dynamic treatment recommendations, although imitation learning limits performance to clinician-level decisions.
 %\cite{peng2021sepsis} combined Deep Q networks with generative and perturbation models to balance imitation and exploration. 
 %, and Kondrup et al. \cite{xia2022safevent} introduced DeepVent, a conservative Q-learning-based agent for ventilator settings. 
\cite{Kuang2024MedReal2Sim} built patient-specific cardiac hemodynamic digital twins via physics-informed self supervised learning.  \cite{lingsch2024fuse} proposed neural surrogates for PDE forward simulation and inverse parameter estimation on simulated data. Most similar to our work, \cite{ogsrl} utilizes a classifier to construct safety constraints by a OOD data classifier for offline RL for Sepsis treatment.  
In contrast to these methods, our work proposes a novel density-based clinically-aware offline learning algorithm, and presents an end-to-end machine learning framework - including a probabilistic digital twin for evaluation, and designing domain-specific medical metrics.

 \vspace{-0.5em}
\section{Background and Formulation}
\label{sec:background}

%In this section, we introduce Offline RL and our real-world clinical application of weaning temporary MCS.

\textbf{Offline Reinforcement Learning.} In this work, we formulate our setting as a Markov decision process (MDP), defined by the tuple $M = (\mathcal{S}, \mathcal{A}, T, r, \mu_0, \gamma)$,
with state space $\mathcal{S}$, action space $\mathcal{A}$, transition dynamics $T: \mathcal{S}\times \mathcal{A} \rightarrow \Delta (\mathcal{S})$, reward function $r(s,a): \mathcal{S}\times \mathcal{A} \rightarrow \R$, initial state distribution $\mu_0$, and discount factor, $\gamma \in (0,1)$.  Reinforcement Learning algorithms aim to find a policy $\pi : \mathcal{S} \rightarrow  \mathcal{A}$ that maximizes the expected cumulative reward
$\mathbb{E}_{\pi, s_0\sim \mu_0} \left[ \sum_{t=0}^{\infty} \gamma^t r(s_t, a_t) \right].$ The optimal policy is defined as, 
\begin{equation}
    \pi^* =\arg\max_{\pi} \mathbb{E}_{\pi, s_0\sim \mu_0} \left[ \sum_{t=0}^{\infty} \gamma^t r(s_t, a_t) \right].
\label{eq:offline_rl}
\end{equation}
%
%The value function 
%\[
%V^{\pi}_{M}(s) := \mathbb{E}_{\pi, T} \left[ \sum_{t=0}^{\infty} %\gamma^t r(s_t, a_t) \,\Big|\, s_0 = s \right]
%\]
%gives the expected discounted return under \( \pi \) when starting from state \( s \).
%
The \textit{Offline RL setting} is when the algorithm only has access to a dataset sampled from the environment $\Denv = \{(s_i, a_i, r_i, s'_i)\}_i$ under a behavior policy $\pi^{\text{B}}$ but cannot interact with the environment.

%We base our solution on the model-based RL algorithms we will have a dynamics model $\hat{T}$  and reward model $\hat{r}$ estimated from the transitions in $\Denv$ via supervised learning. In model-based policy optimization (MBPO) \citep{janner2019trustMBPO}, the algorithm then generates synthetic  $k$-step rollouts under $\hat{T}$ to create a model buffer $\mathcal{D}_{\text{model}}$, and optimizes the policy $\pi(a \mid s)$ on the combined data $ \Denv \cup \mathcal{D}_{\text{model}}$.

%\vspace{-0.5em}
\paragraph{Mechanical Circulatory Support (MCS).}
Left sided forward flow MCS devices are medical devices designed to assist the heart in pumping blood from the left ventricle into the ascending aorta to deliver oxygenated blood to the body. 
%They ensure that blood continues to circulate properly throughout the body when the heart fails to perform this function adequately. 
Cardiogenic Shock (CGS) is a syndrome characterized by cardiac output insufficient for end organ perfusion. Hemodynamically, patients in CGS exhibit low systolic blood pressures, low mean aortic blood pressures, and high heart rates. CGS's mortality rate is historically 50-80\% \citep{vahdatpour2019cardiogenic, sieweke2020mortality}. For patients in severe CGS, MCS plays an integral role in improving blood pressure, %left ventricular unloading, 
maintaining organ perfusion, and aiding heart muscle recovery. As the patient shows signs of improvement, the care team begins to wean the patient from MCS support. The weaning process includes step-wise reduction in MCS performance P-Level with regular assessment of patient response, see Figure \ref{fig:wean_example} for examples. 

%Severity of CGS is graded by the SCAI Shock Criteria for Classification of CS Severity. 

\begin{figure}[h!]
    \centering
    \includegraphics[width=0.9\linewidth]{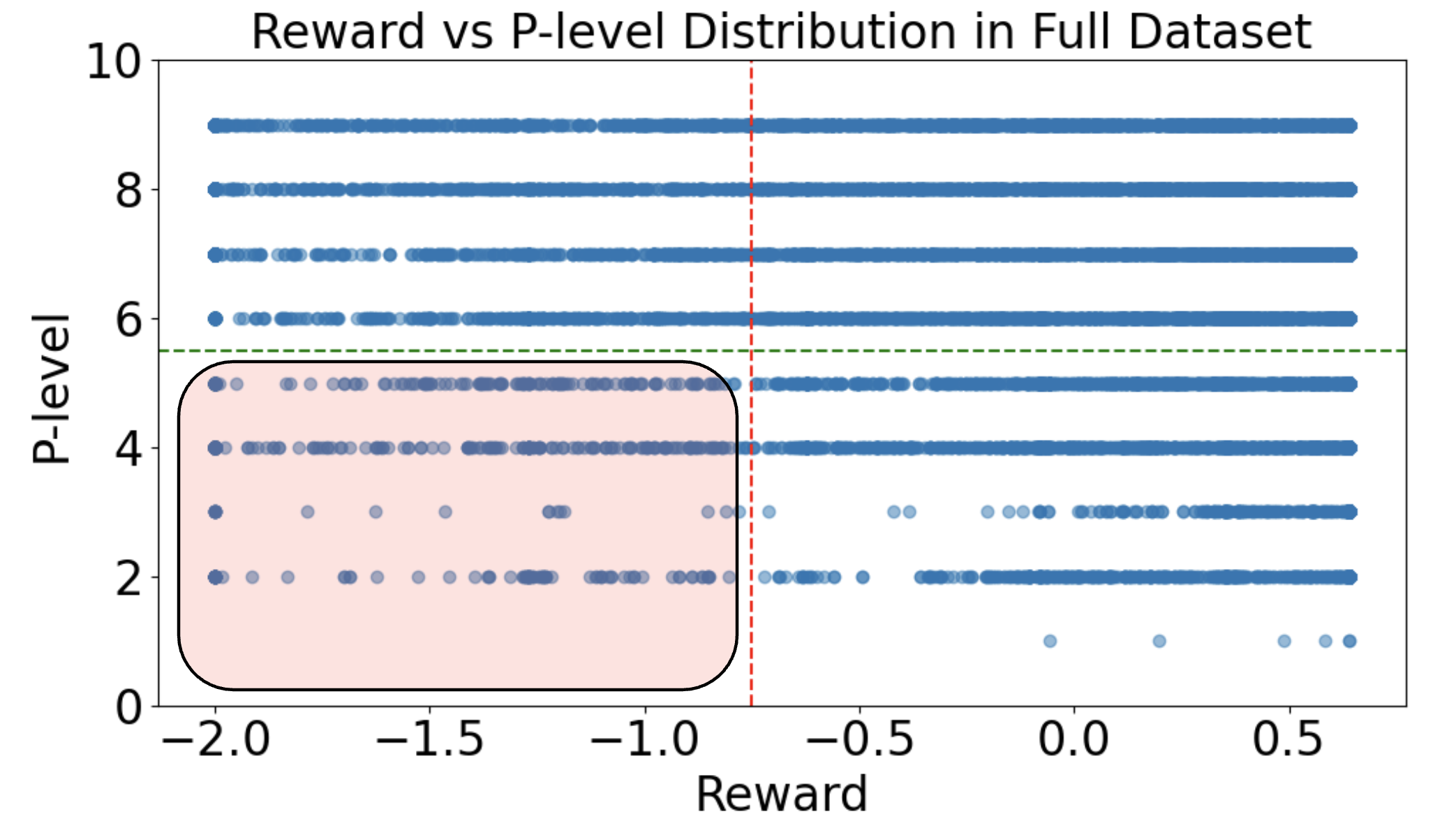}
    \caption{ Illustration of data sparsity in low reward, low P-Level region (shaded in red).}
    \label{fig:sparcity}
    \vspace{-1em}
\end{figure}

%To observe patient response, the clinician must reduce P-level and induce a change in patient state.  
In order to learn and evaluate weaning strategies, we need an environment that predicts the patient response to the proposed change in P-level and evaluates the quality of that P-level choice. 
Although there exist some physics-informed models and numerical simulators \citep{Kuang2024MedReal2Sim, lingsch2024fuse, harviAcademy} of patient hemodynamics, they are often deterministic and not suitable for long time-horizon simulation. Existing solutions fail to account for noise in the real-life patient data and partial observation, due to unobserved treatments (e.g., surgery, medications) and per-patient variability. Figure \ref{fig:sparcity} showcases one of the challenges: data is sparse for ``bad'' cases, where the pump level is low and the patient is unstable (red shaded area). This sparsity results in high epistemic uncertainty in this state-action region. To realistically evaluate a weaning strategy, a digital twin model that can probabilistically quantify uncertainty over a significant time-horizon is integral. 

% \paragraph{Challenges and Desiderata}

% % \begin{itemize} 
% % \item Sparsity of action change penalty metric causes limited penalization on reward. 
% % \item  Performance of our algorithm largely depends on hyperparameter selection of the reward shaping objective. 
% % \item Selecting a threshold to identify OOD samples introduces sensitivity to this hyperparameter. While threshold-based methods can yield high performance, formulating density-based penalization without relying on a threshold can reduce hyperparameter dependence and improve robustness.  
% % \end{itemize}

% Our task has the following desiderata: (1) follow suggested usage (2) clinically salient (3) maximize patient outcome. 

% \sofi{TODO}
% \ays{figure of bad cases missing}

\vspace{-0.5em}
\section{Methodology}
\label{sec:methodology}
In this section, we detail all components of our algorithm (see Appendix \ref{sec:algo} for pseudocode).
\paragraph{MDP Design for MCS.} We first formulate the MCS weaning problem as an MDP. The challenge in formulating the environment is balancing the rich information of medical time series with the learning challenges of a high-dimensional Markov Decision Process (MDP). 
We define each \textit{state} in the MDP to consist of $t$ time-steps of $k$ different physiological features, 
%calculated from the aortic and left ventricular pressure signals recorded by the MCS device, 
i.e. $\mathcal{S} \subseteq \mathbb{R}^{txk}$. The \textit{action} space is $\gA = \{2,3,\cdots, 9\}$, corresponding to pump level P2 to P9 on the MCS device. The objective is to optimize patient outcome with a clinically appropriate weaning strategy. For the offline RL problem, we organize the patient data into a replay buffer dataset of $\D = \{(s_i,a_i, s'_i, r_i)\}_i$  according to the formulation. The state space, action space, reward, and MDP design is informed by expert recommendation and empirical results as presented in Appendix \ref{sec:rl_env}. Under the MDP formulation, we will then describe our digital twin for evaluation, followed by the setup for offline RL training.
\begin{figure*}[h]
    \centering
    \includegraphics[width =0.9\textwidth]{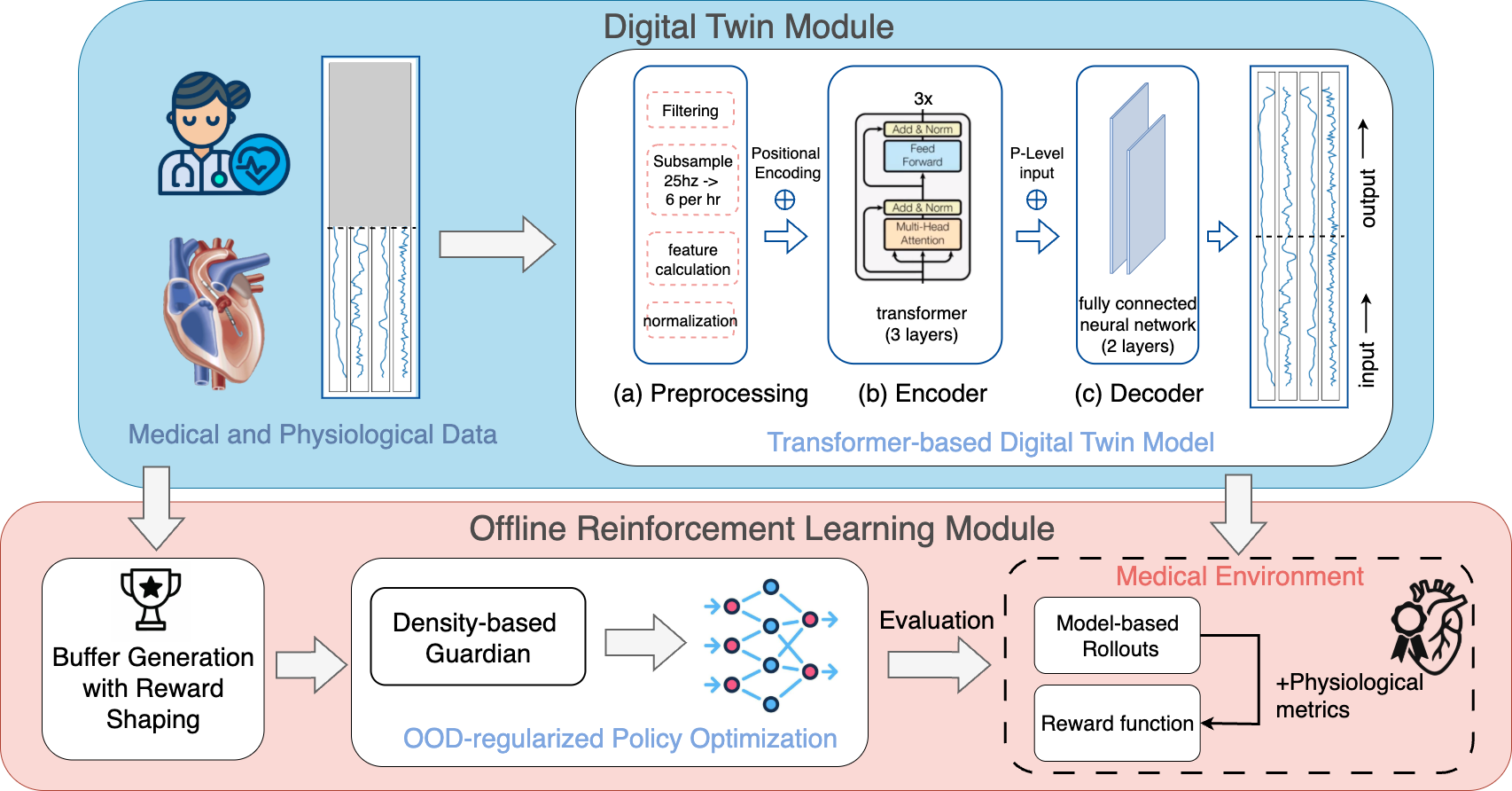}
    \caption{System diagram of the proposed  framework. \textbf{Digital twin module: } We use a transformer encoder to learn a latent representation of the patient’s history, concatenate the representation with the P-Level input, and then decode the output using a fully connected neural network. \textbf{Offline RL module with \textsf{CORMPO}: }  The replay buffer is created from data with clinical guided reward shaping. We learn a density-based guardian model on the data, whose OOD penalty terms are incorporated during policy training. The learned policies are  evaluated in the digital twin-supported medical environment with rich medical metrics.
    }
    \label{fig:system}
    \vspace{-1em}
\end{figure*}

\vspace{-1em}
\subsection{Transformer-based Digital Twin.}
\label{sec:simulator}
\vspace{-0.5em}
To simulate patient trajectories during weaning, we develop a Transformer-based digital twin (TDT) that models patient hemodynamic signals under MCS. The digital twin is denoted as $\hat{T}: \mathcal{S}\times \mathcal{A} \rightarrow \Delta (\mathcal{S})$, which serves as a proxy of the stochastic transition function for the RL task, i.e. $\hat{T}(s,a) = p(s' | s, a)$. At each timestep, the digital twin receives the current patient state, %represented by the 12 clinical variables in the observation space, % —mean arterial pressure (MAP), heart rate, pulsatility, left ventricular relaxation time constant (Tau\_LV), and pump flow— 
along with the control action (for the following time step). %corresponding to the pump motor speed (P-Level) for the following state. 
The digital twin forecasts the next physiological state, enabling safe synthetic ``what-if'' scenarios by simulating patient responses to candidate weaning actions.

The digital twin's model architecture is shown in Figure \ref{fig:system}. The encoder with three multi-head self-attention layers captures temporal dependencies in the multivariate physiological time series. The action is then concatenated to the latent representation and passed to the decoder. The decoder (2 fully connected perceptron layers) predicts the next  state. % enabling safe synthetic ``what-if'' scenarios by simulating patient responses to candidate weaning actions. 
Probabilistic prediction is achieved by retaining dropout ($p=0.1$) in the decoder layers. We train the model to minimize the MSE between the predicted and observed future states, using historical data.

%We use Monte-Carlo Dropout (MC dropout) during training and inference to model uncertainty in our digital twin. Dropout in deep neural networks can be viewed as approximate Bayesian inference \cite{gal2016dropout} and is widely used to quantify uncertainty in data-driven dynamics models \cite{zhang2019quantifying, laves2019well}. During training, dropout is applied after each multi-head self-attention layer of the encoder and feed-forward layers of the decoder, with a fixed probability $p=0.1$. 
%At test time, we retain these dropout masks and perform Monte-Carlo sampling by running $M=50$ stochastic forward passes for each input. Specifically, given a state-action pair $(s_i, a_i)$, we generate a set of stochastic predictions $ s'^{(j)}_{i} = \hat{T}(s_i, a_i), j = 1, \dots, M$, each corresponding to a different stochastic forward pass.

%Concretely, after each multi‑head self‑attention and feed‑forward block we apply a dropout layer with probability $p$, and we retain these dropout masks during inference. Given state $s_{t-1} \text{ and action } a_t$, $\mathbf{s}^{(i)}_t = f_{w}(s_{t-1}, a_t)$, is computed in each stochastic forward pass $i=1,...,M$. 

\vspace{-1em}
\subsection{Offline Reinforcement Learning.}
\label{sec:reward}
%To scale offline RL to the digital twin, We collect the offline patient trajectories into a replay buffer aligned with the state and reward design of the digital twin. 
We propose an algorithm named \textbf{C}linically-aware \textbf{O}OD-regularized \textbf{M}odel-based \textbf{P}olicy \textbf{O}ptimization (\textsf{CORMPO}) for learning clinically informed MCS weaning strategies. Our algorithmic contribution are in two-fold: (1) we designed task-specific reward shaping to incorporate clinical guidelines into the optimization process, and (2) we introduce a density-based OOD-data suppression algorithm to tackle the safety problem in offline reinforcement learning. 

%\vspace{-1em}
\subsubsection{Clinical metrics and reward shaping}

Physiological reward reflects well-being from mean arterial pressure (MAP), heart rate, and pulsatility over the past hour. As we not only value high physiological reward but also gradual P-level changes that lead to stable weaning, we shape the physiological reward with Action Change Penalty (ACP) (similar to \cite{ogsrl}), and Weaning Score (WS). ACP accumulates the magnitude of P-level changes over an episode of length $T$ as
\vspace{-0.5em}
\begin{equation*}
    \text{ACP =} \sum^{T}_{i=1}||a_{i-1}-a_{i}||_2, \text{ if }  ||a_{i-1}-a_{i}||_2 >2.
\end{equation*}
\vspace{-0.25em}
WS rewards the decrease and penalizes the increase in P-level conditioned on the hemodynamic stability of the patient state as follows:
\vspace{-0.25em}
\begin{equation*}
    \text{WS} = \frac{\sum_{i=1}^{T} \mathbb{I}(\texttt{Is\_Stable}(i),1) \cdot \texttt{Weaned}(i)}
    {\sum_{i=1}^{T}  \mathbb{I}(\texttt{Is\_Stable}(i),1)},
    \label{eq:WS}
\end{equation*}
%\begin{equation*}
%\text{Weaned}(i) =
%\begin{cases} 
%-1, & \text{if } a_{i+1} - a_{i} > 0, \\
%a_{i+1} - a_{i}, & \text{if } a_{i+1} - a_{i} \in \{1,2\}, \\
%0, &  \text {otherwise}. \\
%\end{cases}
%\end{equation*}
The definitions for \texttt{Is\_Stable} and  \texttt{Weaned} are in  Appendix \ref{sec:A}.
Finally, we formulate our reward as
\begin{equation}
    r(\cdot) = r_{\text{phys}}(\cdot) - \lambda_1 ACP(\cdot) + \lambda_2 WS(\cdot)  
    \label{shaped rewards}
\end{equation}
where $\lambda_1$, and $\lambda_2$ are hyperparameters determining the magnitude of shaping of each medical metric. 
%We first collect $\Denv$ by organizing patient trajectories into a replay buffer according to our state and reward design. Then, a policy $\hat{\pi}$ can be trained on $\Denv$ with the offline RL objective (Eq. \ref{eq:offline_rl}). 
%

%## Model-Based Offline Reinforcement Learning with Density-Based Uncertainty Quantification

\subsubsection{Density-based guardian for ORL}

%We consider the offline reinforcement learning setting where the agent has access to a static dataset $\mathcal{D}_{\text{env}} = \{(s_i, a_i, r_i, s'_i)\}_{i=1}^N$ collected by a behavior policy $\pi_B$. 
Our goal is to learn a policy $\pi$ that maximizes the expected return in the true MDP $\mathcal{M} = (\mathcal{S}, \mathcal{A}, T, r, \mu_0, \gamma)$, where $T$ denotes the true dynamics, $r$ the reward function, $\mu_0$ the initial state distribution, and $\gamma \in (0,1)$ the discount factor.

Following the model-based approach, we learn a digital twin $\hat{T}$ from $\mathcal{D}_{\text{env}}$, defining the model MDP as $\hat{\mathcal{M}} = (\mathcal{S}, \mathcal{A}, \hat{T}, r, \mu_0, \gamma)$. Let 
\vspace{-0.5em}
\begin{equation}
    \rho^{\pi}_{\hat{T}}(s,a) = \pi(a|s)\sum_{t=0}^{\infty} \gamma^t P^{\pi}_{\hat{T},t}(s)
    \vspace{-0.5em}
\end{equation}
denote the discounted occupancy measure under policy $\pi$ and dynamics $\hat{T}$, where $P^{\pi}_{\hat{T},t}(s)$ is the probability of visiting state $s$ at time $t$. 

We denote the value function under the true dynamics as $V^{\pi}_{\mathcal{M}}(s)$ and the expected return as \[\eta_{\mathcal{M}}(\pi) = \mathbb{E}_{s_0 \sim \mu_0}[V^{\pi}_{\mathcal{M}}(s_0)].\] % We use $\bar{\mathbb{E}}$ to denote improper expectations with respect to $\rho^{\pi}_{\hat{T}}$.

The challenge with model-based offline RL is that learned dynamics models inevitably exhibit varying degrees of accuracy across the state-action space, with errors compounding over multi-step rollouts. While the optimal policy under perfect dynamics may venture beyond the behavioral distribution to achieve higher returns, model inaccuracies in these out-of-distribution (OOD) regions can lead to catastrophic failures, a critical concern in high-stake medical applications. Existing uncertainty-based methods \citep{yu2020mopo, kidambi2020morel} conflates two distinct sources of uncertainty: \textit{aleatoric} uncertainty arising from inherent environment stochasticity, and \textit{epistemic} uncertainty stemming from limited data coverage. This conflation leads to over-penalization of in-distribution (ID) states in noisy environments, unnecessarily constraining the policy in well-understood regions of the state space. Such conservative behavior may prevent the policy from executing known-safe actions that are crucial for task completion. 

As the density of training data determines model reliability, we propose distinguishing ID and OOD states by directly measuring data support through density estimation, which quantifies epistemic uncertainty independent of environment noise. The density-based safeguard enables \textsf{CORMPO} to selectively penalize OOD states and actions while preserving optimal behavior within the data support. %By explicitly quantifying data coverage through KDE, we can construct an uncertainty measure that reflects epistemic uncertainty due to limited data rather than environmental stochasticity, leading to more calibrated conservative behavior.

\paragraph{Density Estimation and Safeguard.} We use kernel density estimation (KDE) in implementation. Given the dataset $\mathcal{D}_{\text{env}}$, the KDE estimator is:
\vspace{-0.5em}
\begin{equation}
 p_{\text{KDE}}(s,a) = \frac{1}{N} \sum_{i=1}^N K_h\left((s,a) - (s_i, a_i)\right)
\end{equation}
where $K_h$ is a kernel function with bandwidth $h$, and $N$ is the number of neighbors closest to each $(s_i, a_i)$.

We define the density regularizer as:
\begin{equation}
u(s,a) = \tau - \log (p_{\text{KDE}}(s,a))
\label{eq:penalty}
\end{equation}
where $\tau$ is the density threshold of in-distribution data. 
%In practice, we use the $5\%$ percentile of $(s,a) \in \Denv$, considering all data above the threshold as ID.
We detail our process of choosing $\tau$ as a percentile in Appendix \ref{sec:imp_det}.
%We upper-bound $u(s,a)$ with interquantile range filtering to avoid extreme penalties.
The density regularizer is then decomposed into $u_+(s,a)=\max\{u(s,a), 0\}$ and  $u_-(s,a)= \min\{u(s,a),0\}$.
%Note that this penalty measure has bounded range: $u(s,a) \in [0, \tau] $. 
Now define the regularized MDP as $\tilde{\mathcal{M}} = (\mathcal{S}, \mathcal{A}, \hat{T}, \tilde{r}, \mu_0, \gamma)$ with:
\begin{equation}
\tilde{r}(s,a) = r(s,a) - \lambda u(s,a)
\label{eq:reg}
\end{equation}
where $\lambda = \gamma c \cdot C_{\hat{T}}$, the symbols will be introduced in the next section.

The optimal policy for the density-penalized MDP is obtained by solving:
\begin{equation}
\hat{\pi} = \arg\max_{\pi} \eta_{\tilde{\mathcal{M}}}(\pi).
\label{eq:final_opt}
\end{equation}

% \begin{algorithm2e}
% \caption{Name}
% \label{alg:algorithm}
%  % older versions of algorithm2e have \dontprintsemicolon instead
%  % of the following:
%  %\DontPrintSemicolon
%  % older versions of algorithm2e have \linesnumbered instead of the
%  % following:
%  \LinesNumbered
% \KwIn{$x_1, \ldots, x_n, w_1, \ldots, w_n$}
% \KwOut{$y$, the net activation}
% $y\leftarrow 0$\;
% \For{$i\leftarrow 1$ \KwTo $n$}{
%   $y \leftarrow y + w_i*x_i$\;
% }
% \end{algorithm2e}

\subsubsection{Theoretical Results}

We show theoretical support that $\hat{\pi}$ as learned by \textsf{CORMPO} has guaranteed performance in the real MDP $\mathcal{M}$, and achieves near-optimal performance among policies that maintain low penalty under the learned dynamics. 

Our guarantee relies on two assumptions. Assumption 1 on bounded rewards is standard in RL theory and holds for most practical applications; assumption 2 captures epistemic uncertainty in supervised learning where model accuracy degrades as we move away from the training distribution.

\begin{assumption}
\textnormal{(Bounded Rewards, Density Regularizer, and Value Functions)}
    The reward function is bounded: $|r(s,a)| \leq r_{\max}$ for all $(s,a) \in \mathcal{S} \times \mathcal{A}$. Consequently, $V^{\pi}_{\mathcal{M}} \in c\mathcal{F}$ where $\mathcal{F} = \{f : \|f\|_{\infty} \leq 1\}$,  $c = r_{\max}/(1-\gamma)$. Let $u: \mathcal S\times\mathcal A\to\mathbb R$ be bounded and write its
    negative part as $u_{-}(s,a)=\min\{u(s,a),0\}\le 0$ with
    $\|u_{-}\|_{\infty} := \sup_{(s,a)} |u_{-}(s,a)| < \infty$. 
        
    %Density regularizer is bounded from below: $||u_-||_\infty = \sup_{(s,a)\in\mathcal S\times\mathcal A}|u_-(s,a)| <\infty$. 
    \label{as:reward}
\end{assumption}

\begin{assumption}\textnormal{(Density-Dependent Model Error) }There exists a constant $C_{\hat{T}} > 0$ such that for all $(s,a) \in \mathcal{S} \times \mathcal{A}$:
\begin{equation}
d_{\mathcal{F}}(\hat{T}(s,a), T(s,a)) \leq C_{\hat{T}} \cdot u_+(s,a) + \epsilon_{\text{approx}}
\end{equation}
where $d_{\mathcal{F}}$ is the integral probability metric w.r.t. $\mathcal{F}$, and $\epsilon_{\text{approx}} > 0$ represents an irreducible approximation error.
    \label{as:error}
\end{assumption} 

%The term $u_-(s,a)$ is the density regularizer in \equationref{eq:penalty}, which reduces the reward for the OOD region.
%measures the ``distance'' from high-density regions. 
The constant $C_{\hat{T}}$ depends on the model class capacity and the smoothness of the true dynamics, which we further discuss in Appendix \ref{sec:proofs}.

Next, define $|G^{\pi}_{\hat{\mathcal{M}}}(s,a)|$ as the difference between the expected value of transition functions $T$ and $\hat{T}$. By the Telescoping Lemma in Appendix \ref{sec:proofs}, we have
\begin{align*}
|G^{\pi}_{\hat{\mathcal{M}}}(s,a)| &= \left|\mathbb{E}_{s' \sim \hat{T}(s,a)}[V^{\pi}_{\mathcal{M}}(s')] - \mathbb{E}_{s' \sim T(s,a)}[V^{\pi}_{\mathcal{M}}(s')]\right| \\
&\leq  c \cdot d_{\mathcal{F}}(\hat{T}(s,a), T(s,a))  \quad \quad \text{(Lemma \ref{lemma:telescoping})}\\
&\leq  c \cdot  C_{\hat{T}} \cdot (u_+(s,a) + \epsilon_{approx}) \; \; \;\text{(Asm. \ref{as:error})}
\end{align*}
where $c$ is specified as in assumption \ref{as:reward}. Bounding value differences allows us to achieve the value bound and optimality results as in uncertainty penalization-based methods \citep{yu2020mopo}:

\begin{mytheorem}\textnormal{(Conservative Value Bound)}
    %Define $u_{sup} = \sup_{\pi} \mathbb{E}_{(s,a) \sim \rho^{\pi}_{\hat{T}}}[u(s,a)]$. 
    Under Assumptions 1-2, for any policy $\pi$:
\begin{equation}
\eta_{\mathcal{M}}(\pi) \geq \eta_{\bar{\mathcal{M}}}(\pi) - \frac{\gamma c \epsilon_{\text{approx}} + \beta}{1-\gamma}
\end{equation}
where $\beta = \lambda\mathbb{E}_{(s,a)\sim\rho_{\hat{T}}^{\pi}} [|u_-(s,a)|]$.
\end{mytheorem}

\begin{mytheorem}\textnormal{(Optimality Gap)} Let $\pi^*$ be the optimal policy for the true MDP $\mathcal{M}$ and $\hat{\pi}$ be the solution to \equationref{eq:final_opt}. Define $\delta_{\min} = \min_{\pi} \mathbb{E}_{(s,a) \sim \rho^{\pi}_{\hat{T}}}[u_+(s,a)]$ and $||u_-||_\infty = \sup_{(s,a)\in\mathcal S\times\mathcal A}|u_-(s,a)|$. Then:
\begin{equation}
\begin{aligned}
\eta_{\mathcal{M}}(\hat{\pi}) \ge\ &
\max_{\pi:\, \mathbb{E}_{\rho^{\pi}_{\hat{T}}}[u_+]\le \delta}\,
\eta_{\mathcal{M}}(\pi)
- 2\lambda\delta
- \frac{\gamma c\,\epsilon_{\text{approx}}}{1-\gamma}
\\[1pt]
&\ - \frac{\lambda}{1-\gamma}\,\|u_-\|_\infty
\end{aligned}
\end{equation}

for any $\delta \geq \delta_{\min}$.
\end{mytheorem}

This result shows that $\hat{\pi}$ achieves near-optimal performance among policies that maintain high density under the learned dynamics. The term $2\lambda\delta$ is the price of conservativeness: the smaller the density budget
$\delta$, the tighter the guarantee but the less exploration. Third term on the RHS reflects irreducible model error in the learned dynamics.
$||u_-||_\infty$ in the last term indicates the maximum value of high-density bonus. So, the last term upper-bounds any optimism induced by rewarding high-density areas.
Intuitively, our density-based penalty implements an adaptive exploration-exploitation trade-off: In high-density regions where $\log (p_{\text{KDE}}(s,a)) \geq \tau$, the term boosts rewards with a positive bonus, allowing the policy to fully exploit the learned model, and in low-density regions where $\log (p_{\text{KDE}}(s,a))$ is close to zero, the penalty discourages exploration into unsafe areas.

% \textbf{Computational Considerations.} The KDE evaluation can be accelerated using:
% \begin{itemize}
% \item Tree-based methods for nearest neighbor search: $O(\log N)$ query time
% \item Random Fourier features for kernel approximation: $O(D)$ evaluation with $D$ features
% \item Caching KDE values for frequently visited states during rollouts
% \end{itemize}

%The uncertainty-penalized MDP framework naturally extends to continuous control tasks where KDE provides smooth uncertainty landscapes, guiding the policy to gradually explore beyond the data support while maintaining safety guarantees.

 \vspace{-0.5em}
\section{Experiments}

\begin{table*}[tbh]
\centering
\resizebox{\textwidth}{!}{
\begin{tabular}{l|ccccc|c}
\toprule
Metric & Expert & BC & \texttt{MBPO} & \texttt{MOPO} & \texttt{SVR}& \textbf{\textsf{CORMPO}} \\
\midrule
\textbf{Phys. Reward ($\uparrow$)} & 1.167 & $0.101 \pm 0.154$ & $1.108 \pm 0.028$ & $1.059 \pm 0.113$ & $0.278 \pm 0.158$ & \textbf{1.224 $\pm$ 0.105} \\
\textbf{ACP ($\downarrow$)}   & 9.26  & $3.703 \pm 0.141$ & $1.963 \pm 0.078$ & $1.907 \pm 0.069$ & $2.945 \pm 0.107$ & \textbf{0.285 $\pm$ 0.035} \\
\textbf{WS ($\uparrow$)}     & -0.091 & $-0.023 \pm 0.008$ & $-0.076 \pm 0.004$ & $-0.021 \pm 0.010$ & $-0.043 \pm 0.007$ & \textbf{0.040 $\pm$ 0.010} \\
\bottomrule
\end{tabular}}
\caption{Trained on the noiseless synthetic dataset, we compare \textsf{CORMPO} against different offline RL models in terms of physiological reward, ACP, and WS ($\uparrow$:  higher is better;  $\downarrow $: lower is better). Evaluation is done on 1000 episodes and averaged over 5 seeds. \textsf{CORMPO} significantly outperforms baselines on physiological reward (14.9\%), ACP (82.6\%), and WS (2.9 times larger than \texttt{MOPO}).}
\label{tab:syn_1}
\end{table*}

\subsection{Experiment Setup}
In this section, we start by introducing the details of our real-life and synthetic datasets. Then, we present results of digital twin learning and the policy evaluation process. Lastly, we compare our proposed model against state-of-the-art offline RL algorithms with qualitative and quantitative results. 
\paragraph{Dataset.}
Our real-life MCS dataset includes 379 patients, with an average length of record of 65.5 hours. We split the patients by ratio 65-15-20 into training, validation, and testing sets. Our clinical data includes 12 features recorded directly or derived from from signals of the MCS device, namely: Mean aortic pressure (MAP), mean pump speed, mean motor current, mean pump flow, left Ventricular Pressure (LVP), left ventricular end diastolic pressure (LVEDP), heart rate (HR), Systolic blood pressure (SBP), Diastolic blood pressure (DBP), Pulsatility, Relaxation Constant (Tau\_LV), and elastance estimation (ESE\_LV). We downsample the original signal of 25 Hz into 0.00167 Hz (1 sample per 10 minutes) and extract samples with a sliding window of 1 hour. 
 
%Acknowledging the challenges of this dataset to develop on Offline RL model; stochasticity in clinician P-levels, and epistemic noise originated by MCS device noise, we create noise-controlled synthetic datasets with our Transformer-based Digital Twin and an expert policy trained on our DT environment. Next, we introduce the synthetic dataset generation procedure alongside TDT performance assessment. 

\paragraph{Digital Twin. }
%We develop a probabilistic digital twin (TDT) as a surrogate of the real-life MCS environment to facilitate the evaluation of Offline RL models. 
Adopting model-based offline policy evaluation (OPE), we utilize our TDT as a surrogate of the real-life MCS environment.
We demonstrate that our TDT successfully captures the underlying data dynamics and associated uncertainty in Appendix \ref{sec:dt_exps}, outperforming 
baselines across accuracy and uncertainty calibration metrics by more than 35\%.  Our Transformer model is more accurate in reflecting response to P-level change and more expressive when capturing large changes in patient states, resulting in its higher accuracy.  %Please see appendix \ref{sec:dt_exps} for detailed introduction of metrics, baselines, and additional results. 
%The strong performance of the Transformer-based architecture is demonstrated through higher prediction variability, and more accurate modeling of P-Level change response, as demonstrated in figure \ref{fig:dt_appendix}.

% \begin{figure*}[h!]
%     \centering
%     \includegraphics[width=\linewidth]{figures/wm1.png}    %\includegraphics[width=\linewidth]{figures/wm2.png}
%     \caption{Digital twin prediction visualization compared with baselines. The Transformer model is more accurate in reflecting response to P-level change and more expressive when capturing large changes in patient state, resulting in its higher accuracy as shown in table \ref{tab:world_model_results}.}
%     \label{fig:digital_twin}
% \end{figure*}

\begin{figure}
    \centering
    \includegraphics[width=\linewidth]{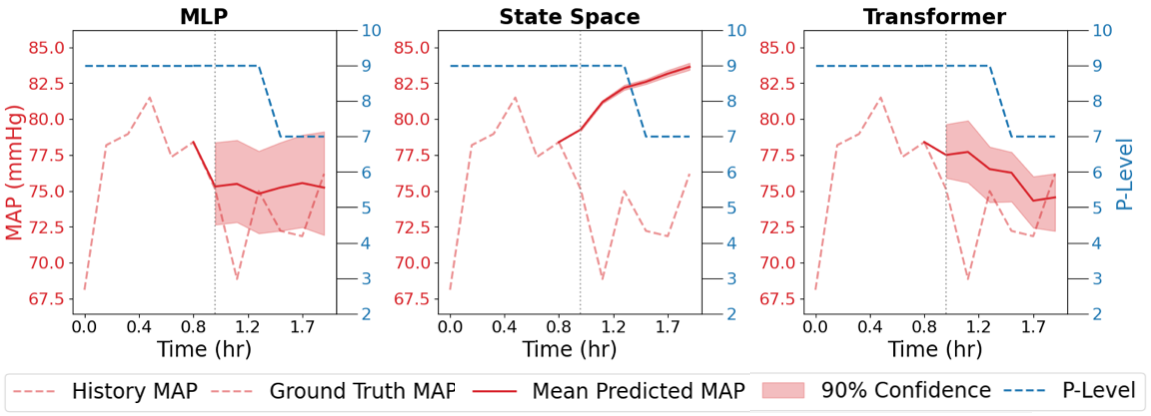}
    \caption{An example of TDT prediction vs. baselines. The Transformer model is more accurate in reflecting response to P-level change and more expressive when capturing large changes in patient state, resulting in its higher accuracy as shown in Table \ref{tab:world_model_results}.}
    \label{fig:wm_small}
    \vspace{-1.5em}
\end{figure}

\paragraph{Synthetic Data Generation.}
We start by constructing a synthetic dataset for a more controlled study of the robustness and efficacy of our algorithm. The synthetic dataset consists of $5000$ trajectory roll-outs from the TDT, starting from a random state in the real dataset and ending at $T=6$, with an expert policy trained with online RL algorithm SAC \citep{haarnoja2018sac}. To mimic the noise setting of our real-life MCS data, we also created a noisy version of the synthetic dataset, by adding a Gaussian noise of $\gN(0, 0.2)$ %0 mean and 0.2 standard deviation
to 80 \% of state transitions. 

\paragraph{Baselines and metrics.}

We compare our proposed method against 4 state-of-the-art Offline RL baselines and the expert policy. For the synthetic experiment, expert is a policy learned in the TDT environment; for the real data experiment, expert is doctor's ground truth actions evaluated with TDT roll-outs. \texttt{Behavioral Cloning (BC)} is a supervised learning baseline trained without exploration. \texttt{Model-Based Policy Optimization (MBPO)} \citep{janner2019trustMBPO} fits a dynamics model and then optimizes the policy on real data augmented with short-horizon rollouts from the dynamics model. \texttt{Model-based Offline Policy Optimization (MOPO)} \citep{yu2020mopo} penalizes the reward using transition uncertainty to encourage conservative policy learning. \texttt{Support Value Regularization (SVR)} \citep{svr} incorporates out-of-distribution (OOD) value regularization to guide policy learning in safe regions.
We evaluate each policy for 1000 episodes of  $T=6$, i.e. a horizon of 6-hour %Evaluation is performed in our simulated medical environment leveraging the TDT as transition dynamics. %TDT predicts the next 1-hour given a randomly sampled observation vector of 1-hour from the train set and a P-level for the next 1-hour. 
and compare performances with respect to physiological reward, Action Change Penalty (ACP), and Weaning Score (WS).

%Gradient-based Weaning Score (WS). We only use GWS for evaluation different from the physiological threshold-based stability definition (see Appendix \ref{sec:A}).\ays{mention why we change it to WS}
%We pick 6 hours as a fixed episode length as it is a reasonable range to observe weaning of an MCS device, considering clinicians practice weaning after observing the hemodynamics state of the patients with MCS implanted for the first 12 hours. 

%We implement reward shaping on MBPO baseline with parameters $\lambda_1$ (ACP weight), and $\lambda_2$ (WS weight). We fit a Gaussian kernel

%\vspace{-0.8em}
\subsection{Synthetic Data Results}
%We train the baselines and \textsf{CORMPO} on the TDT-generated noiseless synthetic dataset of size $30000$. Training and model parameters for the baselines are depicted in Appendix \ref{sec:baselines1}. 

%We implement \textsf{CORMPO} with parameters $\lambda_1$ (ACP weight) as $0.5$ and  $\lambda_2$ (WS weight) as $0.3$. The threshold for log density scores is selected as 5\%. 
We evaluate the performance and noise robustness of \textsf{CORMPO} in the synthetic setting. We present results of learning on the noisy synthetic dataset, deferring implementation details and hyperparameters to Appendix \ref{sec:imp_det}. 

%\begin{figure}
%    \vspace{-1em}
%    \centering
%    \includegraphics[width=0.9\linewidth]{figures/merged_noiseless_svr_cormpo.png}
%    \caption{Comparison of SVR and CORMPO weaning behavior. SVR attains a WS of 1 while CORMPO attains 0 due to the increase in P-level despite patient stability.}
%    \label{fig:placeholder}
 %   \vspace{-1em}
%\end{figure}

In Table \ref{tab:syn_1}, we highlight that our method outperforms baselines in reward and ACP, showcasing the functionality of our reward shaping and MDP design. We demonstrate moderate performance in WS by outperforming all baselines except for \texttt{SVR}.

%\begin{table}[h!]
%  \centering
%  \small
%   \resizebox{\linewidth}{!}{%

%\begin{tabular}{l|ccc|ccc}
%    \toprule
%    \multirow{2}{*}{Model} & \multicolumn{3}{c|}{Reward $\uparrow$}\\  % &\multicolumn{3}{c}{GWS $\uparrow$} 
%    \\
%    & Noiseless & Noisy &\% of Drop \\ %& Noiseless & Noisy & \% of Drop \\
%    \midrule
%    Expert    & 1.5868& 1.4364& 9.48 \\ %&-0.1169& -0.1232& 5.39\\
%    %BC    & 0.479& 1.574 & & -0.05400 & 0.0155 & \\
%    MBPO  &1.773 &  1.590 & 10.32 \\ %& –0.04900 & -0.062 & \textbf{5.08}\\
%    MOPO  & 1.794 & 1.599& 10.87 \\ %& -0.0516 & -0.1513 & 193.4 \\
%    %SVR   & 0.604& 1.328 & & -0.005 & 0.0087 &  \\
%    CORMPO &\textbf{1.828} & \textbf{1.683} &\textbf{7.93}  \\ %& -0.03583& -0.0578&61.3 \\
%    \bottomrule
%  \end{tabular}}
%  \caption{Reward and GWS \% of drop comparison for MBPO, MOPO, and CORMPO trained on noisy and noiseless synthetic datasets. All evaluations are over 100 episodes in the noiseless DT environment.}
%  \label{tab:robustness}
%\end{table}

\begin{table}[h!]
  \centering
  \resizebox{\linewidth}{!}{%
  \begin{tabular}{l|cccc}
    \toprule
\toprule
    Metric & Expert & \texttt{MBPO} & \texttt{MOPO} & \textsf{CORMPO} \\
    \midrule
Reward Noiseless & 1.16 & $1.108 \pm 0.028$ & $1.059 \pm 0.113$ & \textbf{1.224 $\pm$ 0.105} \\
Reward Noisy     & 1.110 & $1.064 \pm 0.031$ & $0.947 \pm 0.094$ & \textbf{1.223 $\pm$ 0.106} \\
\% of Drop       & 5.74 & 3.97 & 10.6 & \textbf{0.082} \\
\bottomrule
  \end{tabular}}
\caption{Comparison of physiological reward under noiseless and noisy settings, and corresponding percentage drop. Evaluation is done on 1000 episodes and averaged over 5 seeds in the noiseless environment setting. \textsf{CORMPO} shows the lowest degradation under noise, indicating robustness.}
\vspace{-0.5em}
\label{tab:robustness}
\end{table}

\begin{table*}[tbh]
\centering
\resizebox{\textwidth}{!}{
\begin{tabular}{l|ccccc|c}
\toprule
Metric & Expert & \texttt{BC} & \texttt{MBPO} & \texttt{MOPO} & \texttt{SVR} & \textbf{\textsf{CORMPO}} \\
\midrule
\textbf{Phys. Reward ($\uparrow$)} & 0.557 & $0.175 \pm 0.118$ & $0.420 \pm 0.139$ & $0.373 \pm 0.129$ & $0.530 \pm 0.152$ & \textbf{0.687 $\pm$ 0.106} \\
\textbf{ACP ($\downarrow$)}    & 1.79  & 0.068 $\pm$ 0.012 & $0.459 \pm 0.032$ & $0.984 \pm 0.020$ & $0.599 \pm 0.057$ & \textbf{0.018 $\pm$ 0.007} \\
\textbf{WS ($\uparrow$)} & 0.053 & \textbf{0.345 $\pm$ 0.008} & $0.147 \pm 0.007$ & $0.042 \pm 0.006$ & $0.166 \pm 0.003$ &$0.173 \pm 0.007$ \\
\bottomrule
\end{tabular}}
\caption{Trained on the real-life dataset, we compare \textsf{CORMPO} against different offline RL baselines in terms of physiological reward, ACP, and WS  with 1000 episodes averaged over 5 seeds. Evaluation is completed in the noiseless environment setting. $\uparrow$: higher is better; $\downarrow$: lower is better. \textsf{CORMPO} outperforms the baselines by 28\% in physiological reward, and 73\% in ACP.}
\label{tab:results_real}
\end{table*}

\begin{figure*}[tbh]
\vspace{-1em}
    \centering
    \includegraphics[width=\linewidth]{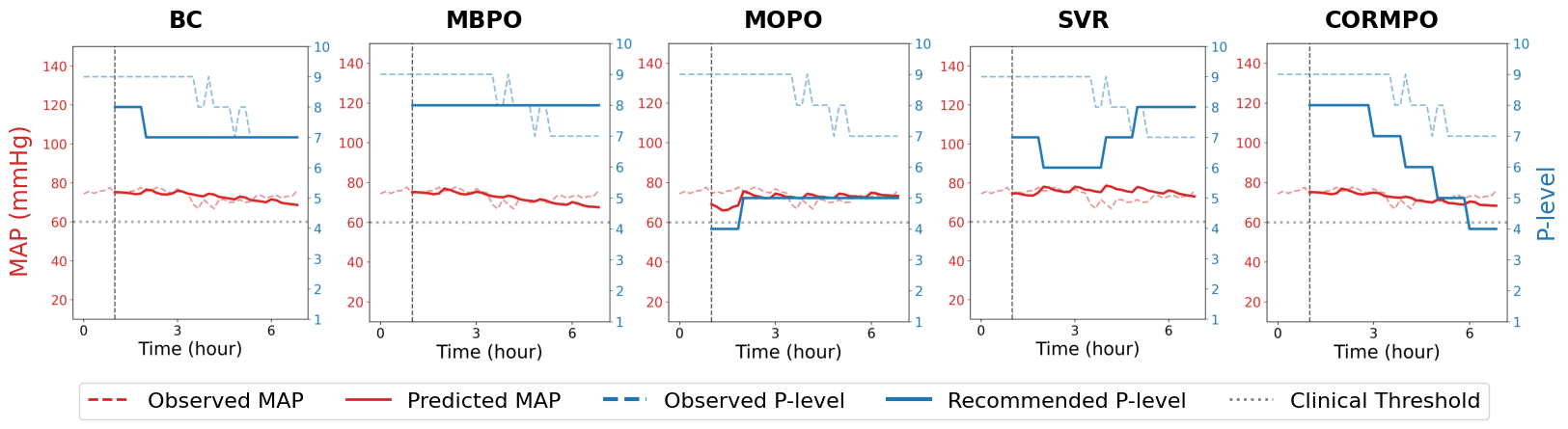}
    \caption{Trained on the real-life dataset, we compare our \textsf{CORMPO} against baselines in 6-hour TDT rollouts. Our TDT predicts stable hemodynamics, away from the clinical threshold, for MAP when guided by \textsf{CORMPO}’s optimal policy. 
    %\textsf{CORMPO} demonstrates successful weaning with $ WS=1.0$ in this specific case. 
    \textsf{CORMPO}'s WS is $1.0$ which is the maximum possible score.
    While \texttt{BC} policy yields limited weaning, \texttt{MBPO}, \texttt{MOPO}, and \texttt{SVR} acts opposite to the weaning behavior because patient stability suggests gradual decrease in P-level.
    %since mean arterial pressure (solid red), heart rate (solid orange), and pulsatility (solid green) are higher than the safe limits indicated in Appendix \ref{sec:A}. 
    %We expect the policies to gradually decrease the P-level %even if the expert P-level does not show weaning. 
    \textsf{CORMPO} results in the most successful weaning in this sample roll-out.}
    % \ays{TODO: get good ACP sample figures}
    \label{fig:real_dataset_figure}
    \vspace{-1em}
\end{figure*}

In Table \ref{tab:robustness}, we emphasize the robustness of \textsf{CORMPO} in reward. 
Notably, it shows the smallest reward degradation under noise, outperforming \texttt{MBPO}, and \texttt{MOPO}.
Taking the expert policy \% of drop in reward as the increase in noise scale, OOD-regularization of \textsf{CORMPO} indicates the functionality of our method. \textsf{CORMPO} also outperforms RL baselines on the noisy synthetic dataset setting. 

\vspace{-0.5em}
 \subsection{Real Dataset Results}
 %\vspace{-0.5em}
% in preamble (optional but nice): \usepackage{caption,booktabs}
%We use the full real-life MCS dataset of 17,865 samples for offline RL training.
%Training and model hyperparameters for the baselines are depicted in Appendix \ref{sec:baselines1}. 
%We implement \textsf{CORMPO} with parameters $\lambda_1$ (ACP weight) as $0.0$ and  $\lambda_2$ (WS weight) as $0.8$. The threshold for log density scores is selected as 5\%.

%\caption{BC (left) and MOPO (right) P-level recommendations (solid blue) and rolled-out digital twin MAP predictions (solid red) for 6 hours compared to the expert P-level and observed MAP.

\begin{figure}[h!]
%\vspace{-1em}
    \centering
    \includegraphics[width=\columnwidth]{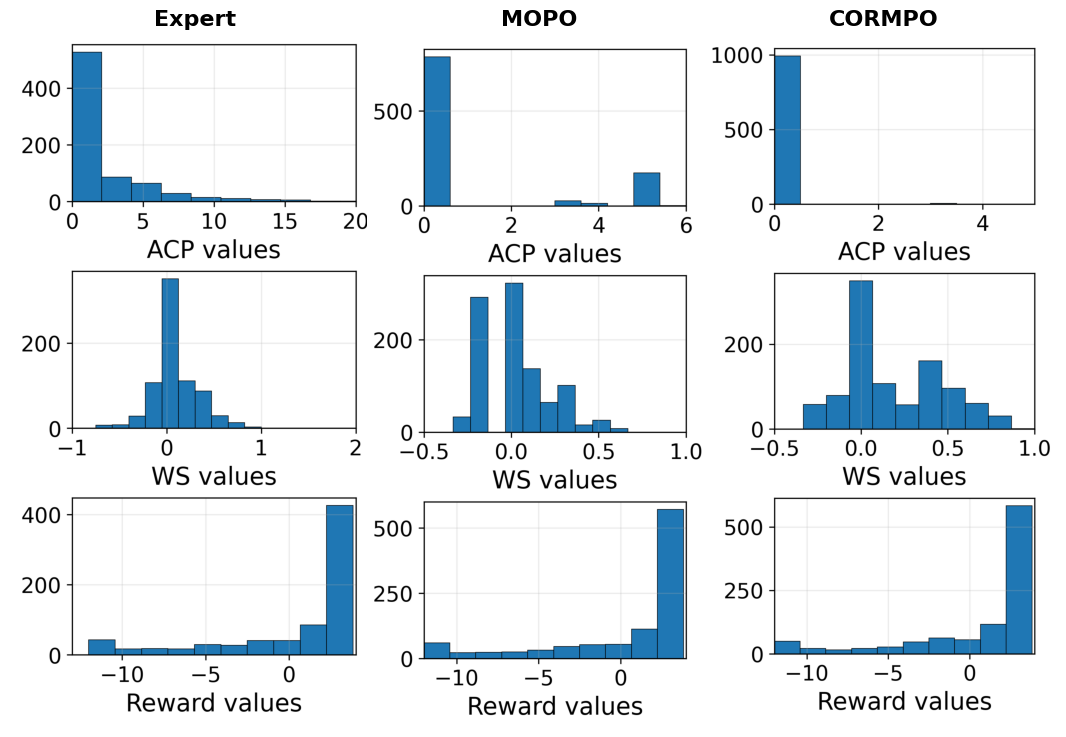}
    \caption{Comparison of physiological reward, ACP, and WS distribution of expert, \texttt{MOPO}, and \textsf{CORMPO} policies. \textsf{CORMPO} suppresses actions with high ACP and results in higher WS compared to baselines, also reducing the high portion of negative rewards in expert policy and achieves higher rewards overall.}
    \label{fig:score_distributions}
    \vspace{-1em}
\end{figure}

We further demonstrate the performance of \textsf{CORMPO} on the real dataset in Figure \ref{fig:real_dataset_figure} and Table \ref{tab:results_real}. In Table \ref{tab:results_real},  \textsf{CORMPO} demonstrates superior overall performance, achieving the highest reward while maintaining the lowest ACP and a WS outperforming most baselines, indicating both successful policy optimization and action stability. 
While \texttt{BC} achieves the highest WS, this is offset by poor reward performance, which can be attributed to the inherent variability in clinician-chosen P-levels and the lack of active exploration leading to over-conservative policy. Specifically, the highest ACP value originating from the clinician P-levels shows the stochasticity of the expert actions, depicted in Figure \ref{fig:score_distributions} row 1 column 1 with the expert resulting in the highest ACP. Therefore, \texttt{BC} avoids P-level increases and defaults to conservative behavior, driven by stochasticity in the clinician data.
\texttt{MBPO} is the third-best performing model after \texttt{SVR}.
The reward of \texttt{MOPO} being worse than \texttt{MBPO} and \textsf{CORMPO} underlines that it over-penalizes the model-based rollouts that diverge into OOD states when the real dataset is already largely stochastic and noisy. 
\texttt{SVR} also demonstrates suboptimal policy performance, as reflected by its high ACP. In contrast, \textsf{CORMPO} pushes the policy in the higher reward region without sacrificing WS and ACP as seen in Figure \ref{fig:score_distributions}. We depict further qualitative and hyperparameter sensitivity analysis of \textsf{CORMPO} in Appendix \ref{sec:additional_results}.
%This highlights the limitations of BC-based OOD regularization and model-free policy optimization approaches in effectively leveraging offline data.

Our KDE implementation is Facebook AI Similarity Search \cite{Johnson2019FaissTBD} which provides a fast approximate nearest-neighbor that has $O(NlogN)$ complexity while traditional KDE’s have $O(N^2)$. Therefore, our KDE training takes 0.37 seconds on our train dataset of 13,399 samples on NVIDIA A100 80GB GPU. We save the model checkpoint and use it for inference during policy training. This does not impose a computational overhead in CORMPO as demonstrated in Table \ref{tab:runtime_comparison}.
\begin{table}[t]
  \captionsetup{width=\columnwidth, justification=raggedright, singlelinecheck=false}
  \centering
  \resizebox{0.9\columnwidth}{!}{%
    \begin{tabular}{l|ccc}
    \toprule
      & \texttt{MBPO} & \texttt{MOPO} & \textbf{\textsf{CORMPO}} \\
    \midrule
      \textbf{Elapsed time (s)} & 1185.670 & 1165.866 & 1195.765 \\
    \bottomrule
    \end{tabular}%
  }
  \caption{\small Elapsed training time of \texttt{MBPO}, \texttt{MOPO}, and \textsf{CORMPO}. Trained for 100 epochs on an NVIDIA A100 80GB  GPU with identical hyperparameters; the small difference ($<\!20$ s) indicates \textsf{CORMPO} adds no significant overhead.}
  \label{tab:runtime_comparison}
  \vspace{-1.5em}
\end{table}

%As a future improvement for CORMPO, we leave the optimal weaning score design to be incorporated in policy optimization and evaluation for future work.  

%In Table \ref{tab:reward_results}, we emphasize that no single offline RL model performs best across all metrics for our task.
%the performance of offline RL policies evaluated with our digital twin.
%For physiological reward, we expect the expert (clinician) policy reward to be close to 0 as it is normalized. Its high ACP and low WS suggest frequent P-level changes despite stable physiology. BC demonstrates good performance in reward and WS, learning from the successful weaning cases.
%MOPO achieves the highest reward 
%supporting uncertainty-penalized rewards in noisy settings,
%although it shows limited weaning based on its low ACP and WS, as in Figure \ref{fig:predict_rl}, indicating a conservative policy. SVR is promising in terms of WS, though it results in lower reward and ACP compared to BC and MOPO, underlining over-regularization. These results indicate the further need for RL models with OOD regularization and uncertainty quantification. 

% \begin{table}[h]
% \centering
% \begin{tabular}{l|l|ccccc}
% \toprule
% Environment & Data type & BC & BCQ & MBPO & MOPO & Ours  \\
% \midrule
% halfcheetah & random & 2.1 \plusminus 0.00 & 2.20 \plusminus 0.02 & 30.7 & 35.4\\
% halfcheetah & expert & 107.0 \plusminus 11.18 & 48.4 \plusminus 0.63 \\
% walker2d & random & 1.6 \plusminus 0.15 & 4.9 \plusminus 0.87 & 8.6 & 13.6 \\
% walker2d & expert & 125.7 \plusminus 6.43 & 86.6 \plusminus 0.92 \\

\vspace{-0.5em}
\section{Discussion}
We presented \textsf{CORMPO}, a comprehensive framework for safe weaning of mechanical circulatory support devices using offline reinforcement learning. Our approach addresses three critical challenges in medical decision-making: the prohibition of online patient interaction, highly uncertain circulatory dynamics, and limited data availability.
We propose a safe offline RL framework that incorporates clinical knowledge via reward shaping and uses density-based regularization to avoid OOD regions, evaluated in our Transformer-based digital twin environment.
%The proposed solution leverages a safe offline RL algorithm that incorporates clinical domain knowledge through reward shaping and integrates density-based regularization on the policy to avoid out-of-distribution areas. To simulate the environment for evaluation of the policies, we integrate a Transformer-based digital twin.
Our theoretical analysis provides performance guarantees under mild assumptions, while experimental validation on synthetic and real patient data demonstrates consistent outperformance of established offline RL baselines across clinically-relevant metrics, including physiological reward, action change penalty, and weaning score. In conclusion, our work presents a complete offline RL methodology for developing a data-driven safe medical decision-making algorithm.

\paragraph{Choice of offline policy evaluation method.} In this paper, we chose model-based offline policy evaluation (OPE) (a widely used approach, see \cite{zhang2021autoregressive, voloshin2019empirical}) over the importance sampling-based approach typically used in medical RL literature \citep{prasad2018weaning}. The choice is primarily because of two reasons: (1) In the medical setting, it is important to show the ``what-if” scenarios resulting from different P-levels to the physicians and decision makers. We developed the digital twin to this end, and use it to evaluate the policies such that it's more interpretable to our medical collaborators. As importance sampling (IS) methods do not provide forecasted trajectories, it makes our system less trustworthy to the physicians. (2) A primary challenge in this problem is sparse data coverage (see Figure \ref{fig:sparcity}). This sparsity fundamentally limits the reliability of Importance Sampling (IS) methods for policy evaluation. Since the dataset consists of human expert actions without an explicit behavior policy, any IS-based approach requires first learning an approximation behavior policy from the data. This introduces a compounding of errors: (a) the inherent high variance of IS in low-coverage regions, where importance weights can become arbitrarily large for state-action pairs rarely visited by the expert, and (b) systematic errors from behavior policy mis-estimation, which are most severe precisely in these same low-coverage regions where we have insufficient data to learn $\pi^{behavior}$ accurately. These two reasons make IS-based evaluation unreliable for our setting. 
Some studies have shown that combining model-based and IS methods result in more robust off-policy evaluation \citep{voloshin2019empirical, thomas2016data}. We defer it to future work to hold a detailed evaluation with the hybrid method. 

%We conclude with a discussion of limitations and future work. 
\paragraph{Limitations and Future Work.} As next steps of this work, we will have the clinical metrics used in this work, such as hemodynamic stability, reviewed by intensive care unit doctors for their suitability and actionability. The definitions are author-designed proxies based on existing device guidance.
How the reward shaping mechanism affects the information quality of the reward signal should be furthered studied as well. 
Additionally, the performance sensitivity to hyperparameter selection for reward shaping and density threshold suggests opportunities for improving the robustness of the algorithm. %emerges us for more stable approaches.
Future work could explore the use of generative density estimators and threshold-free density-based regularization.
For evaluation and applicability, we will explore hybrid OPE methods and transferability of learned policies across different patient populations and clinical settings.

%Our work demonstrates the potential of combining clinical domain expertise with advanced machine learning techniques to address complex medical decision-making problems. Our methodological contributions, particularly the density-based OOD safeguarding and clinically-informed reward design, are applicable to other medical offline RL applications where safety and domain knowledge integration are essential. 
% \vspace{-1em}

\clearpage
%%%%%%%%%%%%%%%%%%%%%%%%%%%%%%%%%%%%%%%%%%%%%%%%%%%%%%%%%%%%
\section*{Acknowledgment}
This work was supported in part by a research grant from Abiomed, Inc., a Johnson \& Johnson company, the U.S. Army Research Office under Army-ECASE award W911NF-07-R-0003-03, the U.S. Department Of Energy, Office of Science, IARPA HAYSTAC Program, and NSF Grants \#2205093, \#2146343, \#2134274, CDC-RFA-FT-23-0069, DARPA AIE FoundSci and DARPA YFA.
\bibliography{jmlr-sample}

\newpage
\appendix

\onecolumn

\section{Medically-informed Metrics}
\label{sec:A}

\textbf{Action Change Penalty (ACP)} \cite{ogsrl}: Abrupt and extreme changes in P-level may maximize rewards; however, they can induce physiological instability in a real-world setting. ACP gauges policy volatility and is given by:
\begin{equation*}
    \text{ACP =}  \sum^{T}_{i=1}||a_{i-1}-a_{i}||_2,
\end{equation*}
where $a_{i-1}$ is an action at state $i-1$, $a_{i}$ is a subsequent action, and $T$ is the episode length. Lower ACP values indicate stable physiology and safe weaning, but note that a value of 0 is undesirable as the P-level must be lowered for weaning. 

%\textbf{Appropriate Intensification Rate (AIR)}: Unstable states in which physiological features are declining require an escalation in treatment. AIR measures the degree of P-level increase after a need for intensification was identified and is given by: 
%\begin{equation*}
%   \text{AIR} = \sum^{T}_{i=1}\text{max(}0, -\bar{\nabla}_i)\times \text{max} (0, a_{i}-a_{i-1}),
%\end{equation*}
% \begin{equation*}
%    \text{AIR} = \sum^{T}_{i=1}\text{ReLU(}-\bar{\nabla}_i)\times \text{ReLU} (a_{i}-a_{i-1}),
% \end{equation*}
%\begin{equation*}
%\bar{\nabla} = \frac{\nabla \text{MAP}(i)+\nabla \text{HR}(i)+\nabla \text{Pulsat}(i)}{3},
%\end{equation*}
%where $\nabla$ is the gradient of the line fitted to a physiological metric over each hour. A higher AIR is desirable and proportional to the product of $-\nabla$ and the action change for a given hour. Low AIR values imply that appropriate actions were rarely taken. \\
\textbf{Weaning Score (WS)}:
To capture satisfactory weaning patterns, we support P-level reductions at most every 1 hour when the patient is observed as hemodynamically stable for the past 1 hour as depicted in Eq. \ref{eq:WS}. Higher weaning scores signify an appropriate reduction in P-level during relatively stable physiological states. We employ 2 definitions of stability based on (1) clinical safety limits and (2) gradient of the past hemodynamic state. First definition is as follows:
%Weaning score measures the quality of removing a patient from MCS if the patient is in hemodynamic stability: 
%\begin{equation*}
%\text{WS} = \sum_{\substack{i=0 \\ \text{MAP}_i > \tau_{\text{MAP}} \\ \text{HR}_i %> \tau_{\text{HR}} \\ \text{Pulsatility}_i > \tau_{\text{Pulsatility}}}}^{T-1}

%\end{equation*}
\begin{align*}
\texttt{Is\_Stable}(i) &= \text{MAP}(i) > \tau_{\text{MAP}} \land\, \text{HR}(i) > \tau_{\text{HR}} \land\, \text{Pulsat}(i) > \tau_{\text{Pulsat}}
\end{align*}
where $\tau_{\text{MAP}}=60, \tau_{\text{HR}}=50$ and $\tau_{\text{Pulsat}}= 10$, indicating limits of hemodynamic stability (see Table \ref{tab:reward} for stability limits). Since our state design represents 1 hour in 10-minute time steps, we calculate the compared MAP value for a state as, $\text{MAP}(i)= \min_{1 \le t \le 6}{\text{MAP}(i,t)}$, same for HR and pulsatility. $T$ is the episode length and $i=0$ indicates the initial state.
Higher weaning scores denote proper lowering of P-level when at a stable state, and low or negative scores imply that P-level is increased despite having healthy physiological indicators. 
Second definition follows analytically as,

\begin{equation}
\begin{aligned}
\texttt{Is\_Stable}(i) &= \left|\frac{\partial\, \mathrm{MAP}(i)}{\partial t}\right| < \tau_1 \land\, \left|\frac{\partial\, \mathrm{HR}(i)}{\partial t}\right| < \tau_2 \land\, \left|\frac{\partial\, \mathrm{Pulsat}(i)}{\partial t}\right| < \tau_3
\end{aligned}
\label{eq:second_ws}
\end{equation}
where $\tau_{\text{MAP}}=1.36, \tau_{\text{HR}}=2.16$ and $\tau_{\text{Pulsat}}= 1.95$, indicating a proxy for stability with a low gradient value of 3 hemodynamic indicators in the past state chosen with statistical significance tests.

Among these two definitions, we utilize the clinical threshold-based stability in the reward shaping while we evaluate the policies with the gradient-based WS definition. Second part of the WS metric is as follows.

%\begin{equation*}
%    \text{WS} = \frac{\sum_{i=0}^{T-1} \mathbb{I}(\text{MAP}(i) > \tau_{\text{MAP}} \land \text{HR}(i) > \tau_{\text{HR}} \land \text{Pulsat}(i) > \tau_{\text{Pulsat}}) \cdot \text{Weaned}(i)}
 %   {\sum_{i=0}^{T-1} \mathbb{I}\big(\text{MAP}(i) > \tau_{\text{MAP}} \land \text{HR}(i) > \tau_{\text{HR}} \land \text{Pulsat}(i) > \tau_{\text{Pulsat}}\big)},
%\end{equation*}
\begin{equation*}
\texttt{Weaned}(i) =
\begin{cases} 
-1, & \text{if } a_{i+1} - a_{i} > 0, \\
a_{i+1} - a_{i}, & \text{if } a_{i+1} - a_{i} \in \{1,2\}, \\
0, &  \text {otherwise}. \\
\end{cases}
\end{equation*}
%\[
%\text{Weaned}(i) \;=\; -\,\mathbb{I}(a_{i+1}-a_i>0)
%\;+\; 0\cdot\mathbb{I}(a_{i+1}-a_i=0)
%\;+\; \mathbb{I}(a_{i+1}-a_i=-1).
%\]

%\textbf{Gradient-Based Weaning Score (GWS)}:
%We offer an alternative method to evaluate weaning strategies in which we support P-level reductions at most every 1 hour when the patient's hemodynamic state is stagnant over for the past 1 hour:

%\begin{equation*}
%    \text{WS} = \frac{\sum_{i=0}^{T-1} \mathbb{I}(|\nabla \text{MAP}(i)| < \tau \land |\nabla\text{HR}(i)| < \tau \land |\nabla\text{Pulsat}(i)| < \tau) \cdot \text{Weaned}(i)}
 %   {\sum_{i=0}^{T-1} \mathbb{I}\big(|\nabla \text{MAP}(i)| < \tau \land |\nabla\text{HR}(i)| < \tau \land |\nabla\text{Pulsat}(i)| < \tau \big)},
%\end{equation*}
%\begin{equation*}
%\text{Weaned}(i) =
%\begin{cases} 
%-1, & \text{if } a_{i+1} - a_{i} > 0, \\
%|a_{i+1} - a_{i}|, & \text{if } a_{i+1} - a_{i} \in \{-1, -2\} \\
%0, &  \text {otherwise}, \\
%\end{cases}
%\end{equation*}

%$\tau = 0.25$.

\textbf{Physiological Reward}: The reward generally reflects the well-being of the patient, according to the mean arterial pressure (MAP), heart rate (HR), and pulsatility of the past hour. Our design follows the clinically defined ranges for hemodynamic stability while caring for the smoothness and differentiability of the function.

\begin{table}[ht]
    \centering
    \begin{tabular}{clll}
    \toprule
    \textbf{Score Component} & \textbf{Value} & \textbf{Score} \\ \midrule
    Hemodynamic \\
    Variable & MAP & $\geq60$ & 0 \\
       & 50 to 59 & 1 \\
      & 40 to 49 & 3 \\
      & $<40$ & 7 \\
    \midrule
      Minimum MAP  & $\geq60$ & 0 \\
      in window & 50 to 59 & 1 \\
      & 40 to 49 & 3 \\
      & $<40$ & 7 \\
     \midrule
      Time Spent MAP& 0 & 0 \\
       $<60$ mmHg (\%)  & 2 & 1 \\
     & & 5 & 3 \\
       & $>5$ & 7 \\
     \midrule
     & Pulsatility & $>20$ & 0 \\
       & 10-20 & 5 \\
       & $<10$ & 7 \\
     \midrule
      HR & $>100$ & 3 \\
       & $<50$ & 3 \\
     \midrule
      LVEDP & $>20$ & 7 \\
       & 15 to 20 & 4 \\
       & $<15$ & 3 \\ 
    \midrule
      CPO & 0.6 to 1 & 1 \\ 
       & $<0.6$ & 3 \\
       & $<0.5$ & 5 \\
     \bottomrule
    \end{tabular}
    \vspace{1em}
     \caption{Hemodynamic instability score table from \cite{buitenwerf2019haemodynamic}. We use a modified version of this table as our physiological reward. When used for evaluating the learned policy as a reward function, we multiply the risk score by -1. }
    \label{tab:reward}
    \end{table}

The reward design in Table \ref{tab:reward} is staircase-shaped, which has two drawbacks:  non-differentiability and a sparse signal. We reformulate the hemodynamic instability score in the following way.

\begin{itemize}
    \item  \textbf{Heart Rate Penalty Function} The heart rate penalty function penalizes deviations from an optimal heart rate of 75 bpm using a quadratic penalty:
\begin{equation}
P_{\text{hr}}(hr) = \text{ReLU}\left(\frac{(hr - 75)^2}{250} - 1\right)
\end{equation}
where $\text{ReLU}(x) = \max(0, x)$. This function has zero penalty for heart rates in the range $[50, 100]$ bpm and applies quadratic penalties for heart rates outside this range.

\item \textbf{Minimum MAP Penalty Function} The minimum Mean Arterial Pressure (MAP) penalty function ensures MAP values remain above 60 mmHg:
\begin{equation}
P_{\text{minMAP}}(MAP) = \text{ReLU}\left(\frac{7(60 - MAP)}{20}\right)
\end{equation}
This function applies a linear penalty when MAP falls below 60 mmHg, with the penalty increasing as MAP decreases further from this threshold.

\item  \textbf{Pulsatility Penalty Function} The pulsatility penalty function maintains pulsatility within the range $[20, 50]$:

\begin{equation}
P_{\text{pulsat}}(p) = \text{ReLU}\left(\frac{7(20 - p)}{20}\right) + \text{ReLU}\left(\frac{p - 50}{20}\right)
\end{equation}

This bi-directional penalty function penalizes pulsatility values below 20 and above 50, with zero penalty for pulsatility in the range $[20, 50]$.

\item \textbf{Hypertension Penalty Function} The hypertension penalty function penalizes elevated mean MAP values above 115 mmHg:

\begin{equation}
P_{\text{hyp}}(MAP) = \text{ReLU}\left(\frac{MAP - 106}{18}\right)
\end{equation}

This function applies a linear penalty for mean MAP values exceeding the hypertension threshold of 106 mmHg.
\end{itemize}

The overall reward function combines all penalty components and negates the sum to create a reward signal:

\begin{equation}
R(s) = -\left[P_{\text{minMAP}}(\min(\text{MAP})) + P_{\text{hyp}}(\overline{\text{MAP}}) + P_{\text{hr}}(\min(HR)) + P_{\text{pulsat}}(\min(\text{Pulsat}))\right]
\end{equation}

where:
\begin{itemize}
    \item $\min(\text{MAP})$, $\min(HR)$, $\min(\text{Pulsat})$ are the minimum values over the time horizon
    \item $\overline{\text{MAP}}$ is the mean MAP over the time horizon
    \item The negative sign converts penalties into rewards (higher rewards for lower penalties)
\end{itemize}

%The function operates on multidimensional time series data with shape $(batch\_size, horizon, features)$, where specific feature dimensions correspond to different physiological parameters.

\section{Markov Decision Process (MDP) Design Details for RL}
\label{sec:rl_env}

\paragraph{Observations. }  The observation space includes 12 hemodynamic features of the patient. Our inputs are the pump pressure, pump speed, and motor current 25 Hz signals recorded by the MCS device. %Impella. 
 We down-sample patient data from 25Hz to 0.00167Hz (1 sample per 10 minutes) and process them into sliding windows of 1 hour (6 time steps) to be used as states for digital twin prediction and decision making based on expert suggestion.
    Therefore, the observation space is $\mathcal{S} = \mathbb{R}^{6 \times 12}$, where each $s_i = x_{t: t+6}$ at some $t$ for a patient. %, for the $t^{th}$ sample. 

\begin{table}[h]
\centering
\resizebox{\textwidth}{!}{
\begin{tabular}{l|*{5}{c}}
\toprule
\toprule
\makecell[l]{In -- out \\ horizon} &
\makecell[c]{15min -- 15min \\ \footnotesize 1 sample / 30s \\ \footnotesize 30ts -> 30ts} &
\makecell[c]{1hr -- 1hr \\ \footnotesize 1 sample / 5min \\ \footnotesize 12ts -> 12ts} &
\makecell[c]{1hr -- 1hr \\ \footnotesize 1 sample / 10min \\ \footnotesize 6ts -> 6ts} &
\makecell[c]{2hr -- 2hr \\ \footnotesize 1 sample / 5min \\ \footnotesize 24ts -> 24ts} &
\makecell[c]{2hr -- 2hr \\ \footnotesize 1 sample / 10min \\ \footnotesize 12ts -> 12ts} \\
\midrule
MSE &
0.234 &
0.142 \plusminus{0.012} &
\textbf{0.124 \plusminus{0.027}} &
0.215 \plusminus{0.009} &
0.159 \plusminus{0.006} \\
MAP MSE &
3.03 &
2.711 \plusminus{0.182} &
\textbf{2.59 \plusminus{0.154}} &
3.583 \plusminus{0.340} &
3.356 \plusminus{0.226} \\
\bottomrule
\bottomrule
\end{tabular}
}
\vspace{1em}
\caption{Alternative settings for the digital twin. Takeaway: shorter horizon and higher down-sampling produce stronger models, but need at least 1 hour of history to provide reasonable action frequency and physiological context.}
\label{tab:alt_setting}
\end{table}
\vspace{-1em}

\paragraph{Action. } The action for our MDP is the pump support level (P-level) of the MCS device. The device operates at 8 different speed levels, from P2-P9, each with a constant motor speed (rpm). The P-level proportionally determines the blood flow provided to the patient by the motor’s speed and current. Clinicians can control the P-level while the patient is on support.
%We use the pump level (p-level) variable, which is the discrete version of pump speed, as the action of our MDP. 
The P-level generally stays unchanged in 1-hour intervals, unlike the state features, since it is manually controlled by the clinicians during the treatment. In practice, we take the mean P-level over the 1-hour interval as expert action. As a result, we define $ \mathcal{A} = \{2, \dots, 9\}$.
    %Since we want to maintain the stationarity of the policy output, i.e., recommended p-levels, in 15 minutes, we preprocess this feature by taking the mean in each 15 minute interval. As a result, we define $a_t \in \{1, \dots,9\}$.
    
\paragraph{Rewards. } The design table for the reward function in Appendix \ref{sec:A} is generated in line with medical consultancy. It assigns a (inverted) risk score based on acceptable intervals for hemodynamic features. 
The physiological reward is further normalized through Z-score normalization and clipped between $[-2,2]$ to ensure training stability. 

\begin{figure}
    \centering
    \includegraphics[width=0.46\linewidth]{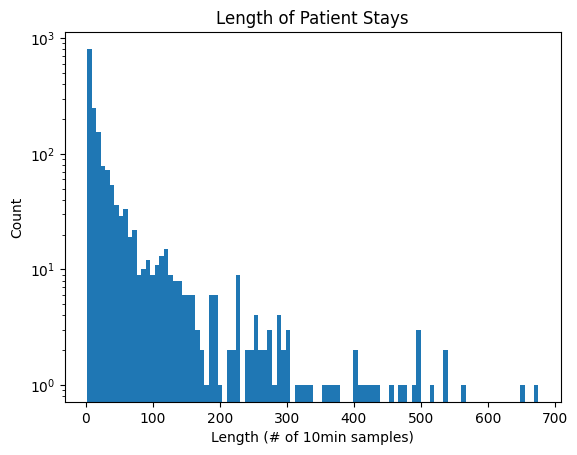}
    \includegraphics[width=0.46\linewidth]{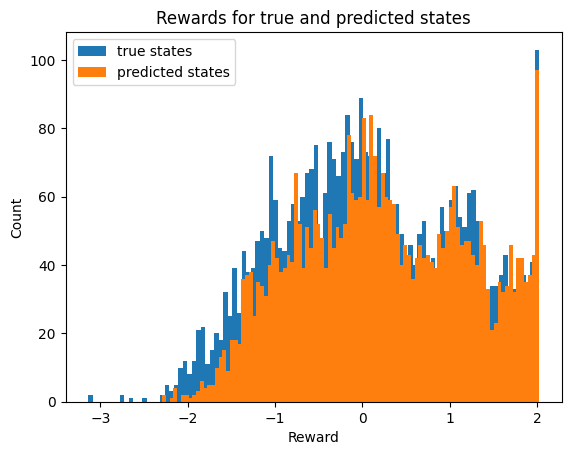}
    \caption{Length distributions of our patient data (left) and the reward score distributions of predicted states versus real patient states.}
    \label{fig:data}
    \vspace{-1em}
\end{figure}

\paragraph{Challenges of Offline RL for MCS} The commonly encountered issue of Offline RL is the limited access to the online environment, which results in distribution shift and large value overestimation errors to account for the shift in the real environment. While these are widely studied problems in RL, medical decision-making introduces other problems: error-prone behavioral policies, highly imbalanced actions in the dataset, and non-differentiable reward functions.  

As there is no golden recipe for weaning a patient from an MCS device, the behavioral policy and the clinician policies are naturally imperfect.
To this end, we expect offline RL to reveal the true policy from the hemodynamic features. Since it is required to simulate the real environment, we largely rely on a digital twin transformer model. However, the model learns to cheat by outputting cardiac cycles copied from the observation distribution. Furthermore, the action space is by definition fully constant in a state, unlike the observation space, which challenges the model compatibility.

\begin{figure}[H]
    \centering
\includegraphics[width=0.48\linewidth]{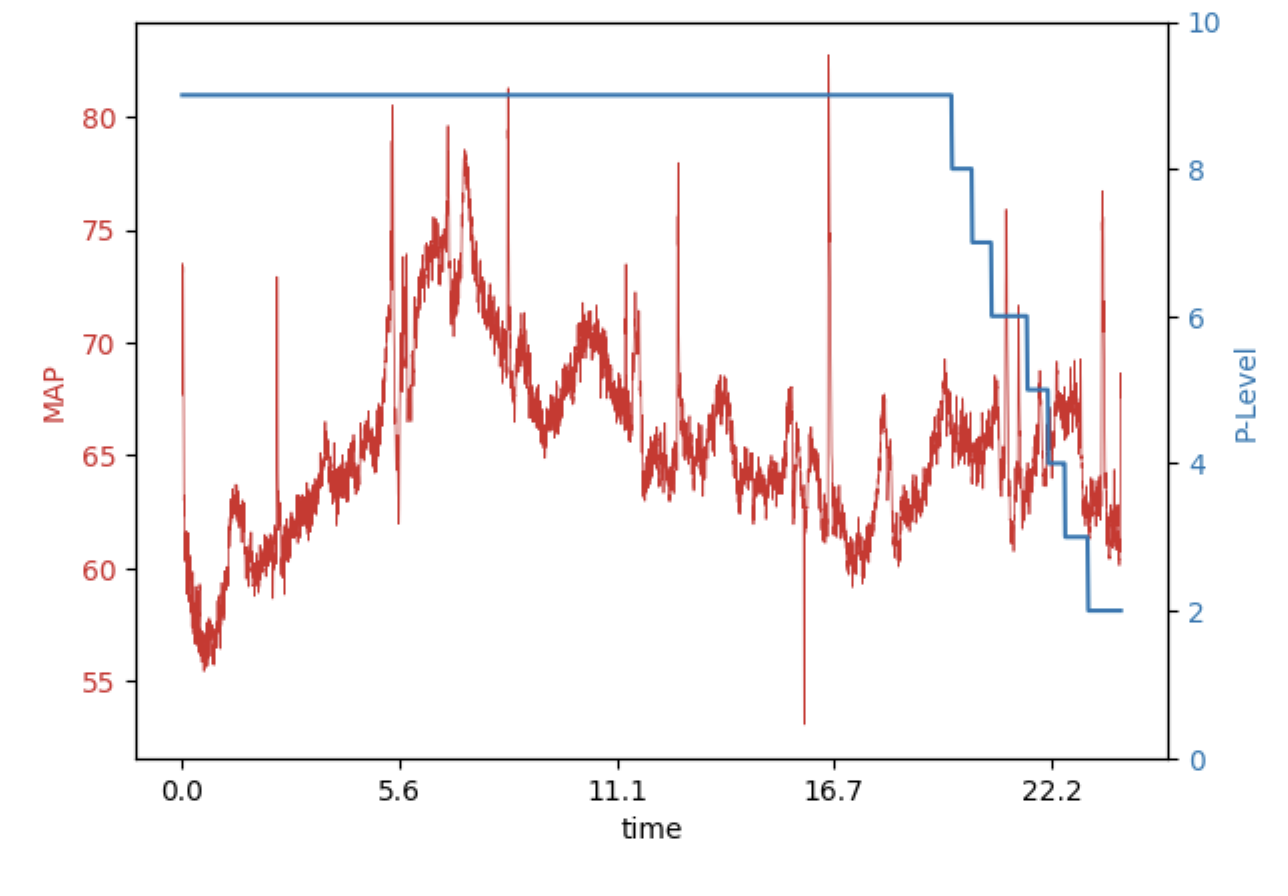}
\includegraphics[width=0.48\linewidth]{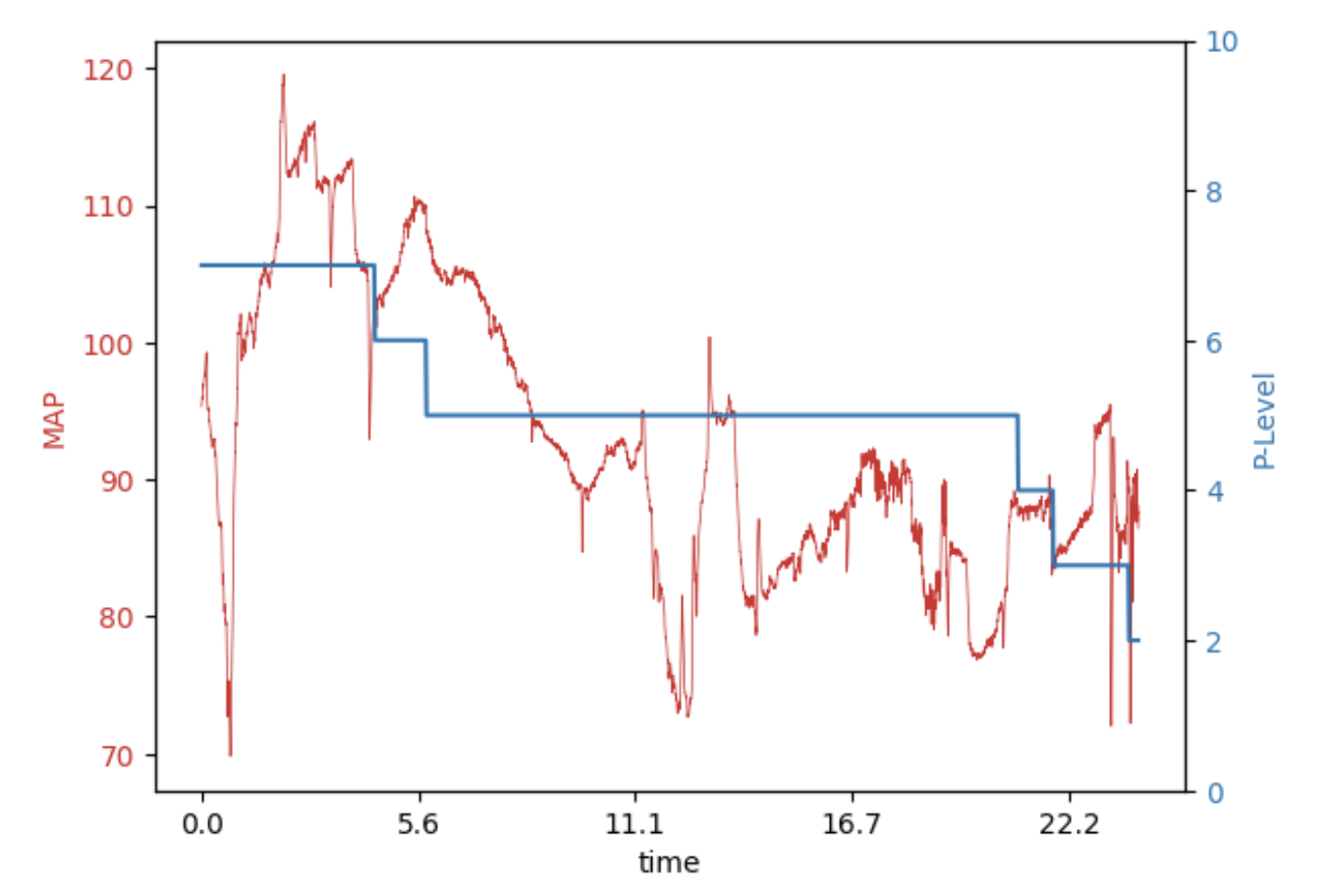}
    \caption{Example weaning of two patients over 24 24-hour horizon.}
    \label{fig:wean_example}
\end{figure}

\paragraph{Example weaning.} We show two examples of doctors' weaning over the course of 24 hours in Figure \ref{fig:wean_example}.

%\newpage
\section{Algorithm Pseudocode}
\label{sec:algo}
\begin{algorithm2e}
\caption{Clinically-aware OOD-regularized Model-based Policy Optimization (CORMPO)}
\label{alg:cormpo}

\KwIn{Dataset $D_{\text{env}} = \{(s_i, a_i, r_i, s'_i)\}_i$, hyperparameters $\lambda_1, \lambda_2, \lambda$, density threshold $\tau$}
\KwOut{Learned policy $\hat{\pi}$}

\tcp{Learn Transformer-based Digital Twin}
Train TDT $\hat{T}: S \times A \rightarrow \Delta(S)$ on $D_{\text{env}}$ using MSE loss with dropout $p = 0.1$\;

\tcp{Learn Density Guardian}
Compute KDE density estimator: $p_{\text{KDE}}(s, a) = \frac{1}{N} \sum_{i=1}^{N} K_h((s, a) - (s_i, a_i))$\;
Define low-density penalty: $u(s, a) = \tau - \log p_{\text{KDE}}(s, a)$\;

\tcp{Create Reward-Shaped Buffer}
Initialize $\gD_{\text{shaped}} \leftarrow \emptyset$\;
\For{each $(s_i, a_i, r_i, s'_i) \in \gD_{\text{env}}$}{
    Compute clinical metrics: $\text{ACP}(a_{i-1}, a_i)$, $\text{WS}(s_{i}, a_{i-1}, a_{i})$\;
    Shape reward: $r(s_i, a_i) \leftarrow r_{\text{phys}}(s_i) - \lambda_1 \text{ACP}(a_{i-1}, a_i) + \lambda_2 \text{WS}(s_{i}, a_{i-1}, a_{i})$\;
    Update buffer: $\gD_{\text{shaped}} \leftarrow \gD_{\text{shaped}} \cup \{(s_i, a_i, \tilde{r}(s_i, a_i), s'_i)\}$\;
}

\tcp{Learn Policy with OOD Regularization}
Define penalized MDP $\tilde{M} = (S, A, \hat{T}, \tilde{r}, \tilde{\mu}_0, \gamma)$ with:\;
$\tilde{r}(s, a) \leftarrow r(s, a) - \lambda u(s, a)$ where $\lambda = \gamma c \cdot C_{\hat{T}}$\;

Train policy $\hat{\pi}$ using model-based RL (MBPO) on $\tilde{M}$\;
$\hat{\pi} \leftarrow \arg\max_{\pi} \eta_{\tilde{M}}(\pi)$\;

\tcp{Evaluate Policy}
Evaluate $\hat{\pi}$ in TDT environment using clinical metrics (Physiological reward, ACP, WS)\;

\Return{$\hat{\pi}$}
\end{algorithm2e}

\section{Proofs and supporting theoretical results}
\label{sec:proofs}
\begin{mylemma}[Telescoping lemma - Lemma 4.1 in \cite{yu2020mopo}]
Let $M$ and $\widehat{M}$ be two MDPs with the same reward function $r$, but different dynamics $T$ and $\widehat{T}$ respectively. Let 
\[
G^{\pi}_{\widehat{M}}(s,a) := \mathbb{E}_{s' \sim \widehat{T}(s,a)}\big[ V^{\pi}_{M}(s') \big] - \mathbb{E}_{s' \sim T(s,a)}\big[ V^{\pi}_{M}(s') \big].
\]

\noindent Then,
\begin{equation}
    \eta_{\widehat{M}}(\pi) - \eta_{M}(\pi) 
    = \gamma \, \mathbb{E}_{(s,a)\sim \rho^{\pi}_{\widehat{T}}} \big[ G^{\pi}_{\widehat{M}}(s,a) \big]
\end{equation}

\noindent As an immediate corollary, we have
\begin{equation}
\begin{aligned}
    \eta_{M}(\pi) 
    &= \mathbb{E}_{(s,a)\sim \rho^{\pi}_{\widehat{T}}} \Big[ r(s,a) - \gamma G^{\pi}_{\widehat{M}}(s,a) \Big] \\
    &\geq \mathbb{E}_{(s,a)\sim \rho^{\pi}_{\widehat{T}}} \Big[ r(s,a) - \gamma \big| G^{\pi}_{\widehat{M}}(s,a) \big| \Big]
\end{aligned}
\end{equation}

Leveraging properties of $V^\pi_\gM$, we will replace $G^{\pi}_\gM$ by an upper bound that depends solely on the error of the
dynamics $T$. We first note that if $\mathcal{F}$ is a set of functions mapping $S$ to $\R$ that contains $V^\pi_\gM$ then,
\begin{equation}
\big| G^{\pi}_{\widehat{M}}(s,a) \big| 
\leq \sup_{f \in \mathcal{F}} \left| 
\mathbb{E}_{s' \sim \widehat{T}(s,a)}[f(s')] 
- \mathbb{E}_{s' \sim T(s,a)}[f(s')] 
\right| 
=: d_{\mathcal{F}}\!\left(\widehat{T}(s,a), T(s,a)\right),
\end{equation}
\label{lemma:telescoping}
\end{mylemma}

\subsection{Proof of Theorem 1.} 

%\begin{align}
%\eta_{\mathcal{M}}(\pi) &= \eta_{\hat{\mathcal{M}}}(\pi) - \gamma \bar{\mathbb{E}}_{(s,a) \sim \rho^{\pi}_{\hat{T}}}[G^{\pi}_{\hat{\mathcal{M}}}(s,a)]\\
%&\geq \eta_{\hat{\mathcal{M}}}(\pi) - \gamma \bar{\mathbb{E}}_{(s,a) \sim \rho^{\pi}_{\hat{T}}}[|G^{\pi}_{\hat{\mathcal{M}}}(s,a)|]\\
%&\geq \eta_{\hat{\mathcal{M}}}(\pi) - \gamma \bar{\mathbb{E}}_{(s,a) \sim \rho^{\pi}_{\hat{T}}}[c \cdot (C_{\text{model}} \cdot u_+(s,a) + \epsilon_0)]\\
%&= \bar{\mathbb{E}}_{(s,a) \sim \rho^{\pi}_{\hat{T}}}[r(s,a)] - \lambda \bar{\mathbb{E}}_{(s,a) \sim \rho^{\pi}_{\hat{T}}}[u_+(s,a)] - \gamma c \epsilon_0 \bar{\mathbb{E}}_{(s,a) \sim \rho^{\pi}_{\hat{T}}}[1]\\
%&= \bar{\mathbb{E}}_{(s,a) \sim \rho^{\pi}_{\hat{T}}}[\tilde{r}(s,a)] - \frac{\gamma c \epsilon_0}{1-\gamma}\\
%&= \eta_{\tilde{\mathcal{M}}}(\pi) - \frac{\gamma c \epsilon_0}{1-\gamma} \qquad %\square
%\end{align}
Starting from Lemma \ref{lemma:telescoping} (Lemma 4.1 in \cite{yu2020mopo}, known as the telescoping lemma):
\begin{equation}
\eta_{\mathcal{M}}(\pi) = \eta_{\hat{\mathcal{M}}}(\pi) - \gamma\mathbb{E}_{(s,a)\sim\rho^{\pi}_{\hat{T}}} [G^\pi_{\hat{\mathcal{M}}}(s,a)] \geq \eta_{\hat{\mathcal{M}}}(\pi) - \gamma\mathbb{E}_{(s,a)\sim\rho^{\pi}_{\hat{T}}} [|G^\pi_{\hat{\mathcal{M}}}(s,a)|] 
\end{equation}
\noindent By Assumption 2:
\begin{equation}
\geq \eta_{\hat{\mathcal{M}}}(\pi) - \gamma c C_{\hat{T}}\mathbb{E}_{(s,a)\sim\rho^{\pi}_{\hat{T}}}[u_+(s,a)] - \gamma c\epsilon_0 
\end{equation}
\noindent Since $\eta_{\hat{\mathcal{M}}}(\pi) = \mathbb{E}_{(s,a)\sim\rho^{\pi}_{\hat{T}}}[r(s,a)]$, add and subtract $\lambda\mathbb{E}_{(s,a)\sim\rho^{\pi}_{\hat{T}}}[u(s,a)]$:
\begin{multline}
= \mathbb{E}_{(s,a)\sim\rho^{\pi}_{\hat{T}}}[\bar{r}(s,a)] + \lambda\mathbb{E}_{(s,a)\sim\rho^{\pi}_{\hat{T}}}[u_+(s,a) + u_-(s,a)] \\ - \gamma c C_{\hat{T}}\mathbb{E}_{(s,a)\sim\rho^{\pi}_{\hat{T}}}[u_+(s,a)] - \gamma c\epsilon_0 
\end{multline}
\noindent Regrouping terms and using $u_-(s,a) \leq 0$:
\begin{equation}
\geq \mathbb{E}_{(s,a)\sim\rho^{\pi}_{\hat{T}}}[\bar{r}(s,a)] + (\lambda - \gamma c C_{\hat{T}})\mathbb{E}_{(s,a)\sim\rho^{\pi}_{\hat{T}}}[u_+(s,a)] - \lambda\mathbb{E}_{(s,a)\sim\rho^{\pi}_{\hat{T}}}[|u_-(s,a)|] - \gamma c\epsilon_0 
\end{equation}
\noindent Summing over the infinite horizon:
\begin{equation}
= \eta_{\bar{\mathcal{M}}}(\pi) - \frac{\gamma c \epsilon_{\text{approx}}}{1-\gamma} - \frac{\lambda - \gamma c C_{\hat{T}}}{1-\gamma}\mathbb{E}_{\rho^{\pi}_{\hat{T}}}[u_+] - \frac{\lambda}{1-\gamma}\mathbb{E}_{\rho^{\pi}_{\hat{T}}}[|u_-|] 
\end{equation}
Using $\lambda = \gamma c C_{\hat{T}}$,
\begin{equation}
= \eta_{\bar{\mathcal{M}}}(\pi) - \frac{\gamma c \epsilon_{\text{approx}}}{1-\gamma} - \frac{\lambda}{1-\gamma}\mathbb{E}_{\rho^{\pi}_{\hat{T}}}[|u_-|]
\end{equation}

\noindent This completes the proof. $\square$

\begin{mylemma}[Discounted bound for signed shaping]
Starting from Assumption \ref{as:reward}, for all $(s,a)$, $u(s,a)\ge u_{-}(s,a)\ge -\|u_{-}\|_{\infty}$.
Thus, for any policy $\pi$, any dynamics, and any $\gamma\in(0,1)$,
\[
\mathbb{E}\!\left[\sum_{t=0}^{\infty}\gamma^{t} u(s_t,a_t)\right]
\;\ge\; -\,\sum_{t=0}^{\infty}\gamma^{t}\|u_{-}\|_{\infty}
\;=\; -\,\frac{\|u_{-}\|_{\infty}}{1-\gamma}.
\]

Consequently, for any $\lambda\ge 0$,
\[
\lambda\,\mathbb{E}\!\left[\sum_{t=0}^{\infty}\gamma^{t}\,u(s_t,a_t)\right]
\;\ge\; -\,\frac{\lambda}{1-\gamma}\,\|u_{-}\|_{\infty}.
\]
\label{lemma:signed}
\end{mylemma}

\subsection{Proof of Theorem 2.}

We first note that a two-sided bound follows from Lemma~\ref{lemma:telescoping} 
and Assumption~\ref{as:error}:
\begin{equation}
\big| \eta_{\widehat{\mathcal M}}(\pi) - \eta_{\mathcal M}(\pi) \big|
\le \gamma c\,\mathbb{E}_{(s,a)\sim\rho_{\widehat T}^{\pi}}
\!\big[ d_{\mathcal F}(\widehat T(s,a), T(s,a)) \big]
\le \gamma c\,C_{\widehat T}\,\mathbb{E}_{(s,a)\sim\rho_{\widehat T}^{\pi}}[u_+(s,a)]
+ \frac{\gamma c}{1-\gamma}\epsilon_{\text{approx}}.
\label{eq:two_sided}
\end{equation}
Next, recall that $\widehat{\pi}$ is optimal in the penalized MDP $\widehat{\mathcal M}$:
\begin{equation}
\widehat{\pi} = \arg\max_{\pi}\, 
\mathbb{E}_{(s,a)\sim\rho_{\widehat T}^{\pi}}\!\big[ r(s,a) - \lambda\,u(s,a) \big].
\label{eq:penalized_policy}
\end{equation}
Thus, for any policy $\pi$,
\begin{align}
\eta_{\mathcal M}(\widehat{\pi}) 
&\ge \eta_{\widehat{\mathcal M}}(\widehat{\pi}) 
- \frac{\gamma c}{1-\gamma}\epsilon_{\text{approx}}
\quad &&\text{(by \eqref{eq:two_sided})}
\\
&\ge \eta_{\widehat{\mathcal M}}(\pi)
- \frac{\gamma c}{1-\gamma}\epsilon_{\text{approx}}
\quad &&\text{(by optimality of $\widehat{\pi}$ in $\widehat{\mathcal M}$)}
\\
&= \mathbb{E}_{(s,a)\sim\rho_{\widehat T}^{\pi}}[r(s,a)]
- \lambda\,\mathbb{E}_{(s,a)\sim\rho_{\widehat T}^{\pi}}[u(s,a)]
- \frac{\gamma c}{1-\gamma}\epsilon_{\text{approx}}
\quad &&\text{(by \eqref{eq:penalized_policy})}
\\
&= \mathbb{E}_{(s,a)\sim\rho_{\widehat T}^{\pi}}[r(s,a)]
- \lambda\,\mathbb{E}_{(s,a)\sim\rho_{\widehat T}^{\pi}}[u_+(s,a)]
- \lambda\,\mathbb{E}_{(s,a)\sim\rho_{\widehat T}^{\pi}}[u_-(s,a)]
- \frac{\gamma c}{1-\gamma}\epsilon_{\text{approx}}
\\
&\ge \mathbb{E}_{(s,a)\sim\rho_{\widehat T}^{\pi}}[r(s,a)]
- \lambda\,\mathbb{E}_{(s,a)\sim\rho_{\widehat T}^{\pi}}[u_+(s,a)]
- \frac{\gamma c}{1-\gamma}\epsilon_{\text{approx}}
- \frac{\lambda}{1-\gamma}\|u_-\|_\infty,
\label{eq:ineq1}
\end{align}
where the last inequality follows from the bounded discounted sum of negative bonuses 
(Lemma~\ref{lemma:signed}), i.e.,
\(
\mathbb{E}\!\big[\sum_{t\ge0}\gamma^t u_-(s_t,a_t)\big] 
\ge - \tfrac{\|u_-\|_\infty}{1-\gamma}.
\)
Finally, using $\mathbb{E}_{(s,a)\sim\rho_{\widehat T}^{\pi}}[u_+(s,a)] \le \delta$ for any 
policy $\pi$ within the model-error budget $\delta$, we obtain:
\begin{equation}
\eta_{\mathcal M}(\widehat{\pi}) 
\ge \max_{\pi:\,\mathbb{E}_{(s,a)\sim\rho_{\widehat T}^{\pi}}[u_+]\le\delta}
\mathbb{E}_{(s,a)\sim\rho_{\widehat T}^{\pi}}[r(s,a)]
- 2\lambda\delta
- \frac{\gamma c}{1-\gamma}\epsilon_{\text{approx}}
- \frac{\lambda}{1-\gamma}\|u_-\|_\infty,
\end{equation}
for all $\delta \ge \delta_{\min}$, which completes the proof.  $\square$

\subsection{Transformer Approximation Bound for Transition Functions}

Let $\pi$ be any feasible solution and $\hat{T}_\theta$ be the transformer-learned transition function with parameters $\theta$. Assume the following:
\begin{itemize}
    \item Dataset $\mathcal{D} = \{(s_i, a_i, r_i, s'_i)\}_{i=1}^N$ with $N$ samples
    \item Transformer with $L$ layers, dimension $d_{model}$, and $H_{attn}$ attention heads
    \item Lipschitz continuous true transition $T$ with constant $L_T$
\end{itemize}

Then, with probability at least $1 - 2\beta - 4\delta$, the following holds:

$$\left|V_{\phi, \hat{T}_\theta}^{\pi}(\rho_0) - V_{\phi, T}^{\pi}(\rho_0)\right| \leq \epsilon_H + \epsilon_{trans}$$

where:

$$\epsilon_H := \frac{\gamma^{H+1}(2-\gamma)\phi_{max}}{(1-\gamma)^2}$$

$$\epsilon_{trans} := \frac{\phi_{max}(\gamma - \gamma^{H+2})}{(1-\gamma)^2} \left( \epsilon_{approx} + \epsilon_{gen} \right)$$

with:

$$\epsilon_{approx} := C_{trans} \cdot \min\left\{\frac{1}{L \cdot d_{model}}, \frac{1}{H_{attn} \cdot N_{ctx}}\right\}$$

$$\epsilon_{gen} := L_T \sqrt{\frac{d \log(1/\delta) + \log N}{N}} + \mathcal{O}\left(\frac{1}{\sqrt{N_{eff}}}\right)$$

Assumption 2 is supported by the above derivation of transformer approximation error.  In regions with high KDE density $p_{\text{KDE}}(s,a)$, the local sample density is high, yielding large $N_{\text{eff}}$ and small error. Conversely, as $p_{\text{KDE}}(s,a) \rightarrow 0$, the effective sample size diminishes, causing the error to grow. The linear relationship $\tau - p_{\text{KDE}}(s,a)$ captures this first-order dependence between local data density and model error, while $\epsilon_{\text{approx}} = C_{\text{trans}} \cdot \min\{1/(L \cdot d_{\text{model}}), 1/(H_{\text{attn}} \cdot N_{\text{ctx}})\}$ represents the transformer's irreducible approximation error determined by its architecture (depth $L$, dimension $d_{\text{model}}$, and attention heads $H_{\text{attn}}$). The constant $C_{\hat{T}}$ encapsulates the Lipschitz constant $L_T$ of the true dynamics and the dimensionality-dependent factors.

% ## Key Components:

% 1. **$\epsilon_H$**: Horizon truncation error (identical to KDE case)

% 2. **$\epsilon_{approx}$**: Transformer's approximation capacity
%    - Decreases with depth $L$ and model dimension $d_{model}$
%    - $N_{ctx}$ is the context length for historical states
%    - $C_{trans}$ depends on smoothness of transition dynamics

% 3. **$\epsilon_{gen}$**: Statistical generalization error
%    - First term: standard learning theory bound
%    - $N_{eff} = N / (1 + \lambda_{rank})$ where $\lambda_{rank}$ measures effective rank of attention matrices

\section{Digital Twin Experiment Details}
\label{sec:dt_exps}
\subsection{Digital Twin Experiments}
\begin{figure}[tbh]
    \centering
    \includegraphics[width=0.95\linewidth]{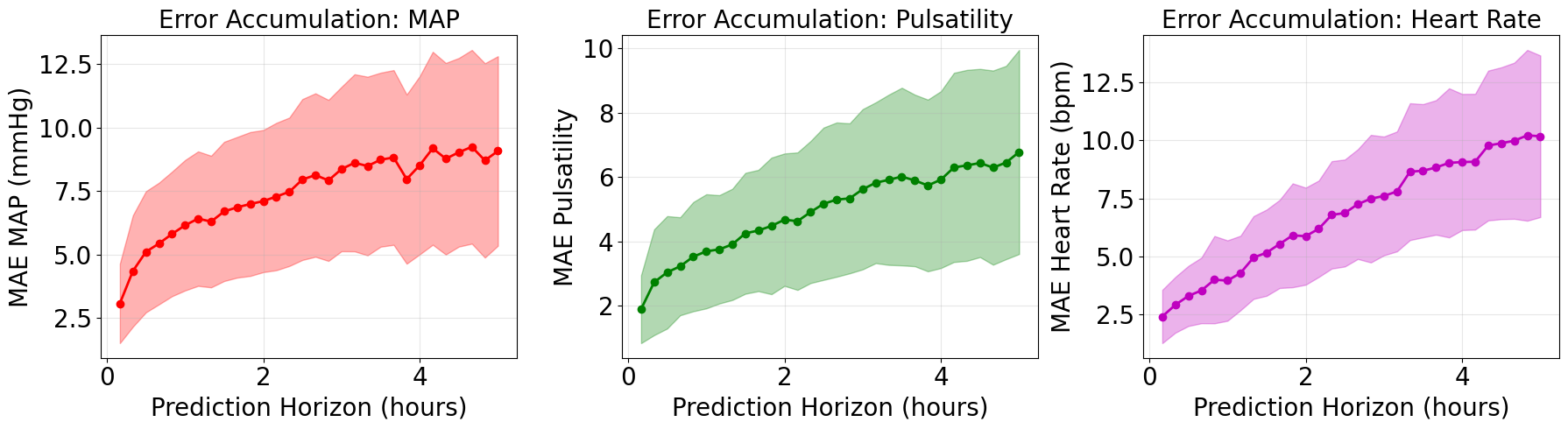}
    \caption{Error accumulation of the digital twin predictions (mean $\pm$ standard deviation evaluated across 500 samples). We can see that the error, especially mean average error of MAP, accumulates slowly through the horizon. }
    \label{fig:error-acc}
\end{figure}

We show experiment results of various digital twin models in Table \ref{tab:world_model_results}.  The transformer based digital twin with sinusoidal positional embeddings outperforms all baselines, including the transformer architecture employing rotary positional embeddings. Figure \ref{fig:error-acc} provides an analysis of the error accumulation over prediction horizon for our digital twin - for the three metrics that we base our reward on, the error accumulates sub-linearly and is low even at the 6 hour horizon. 

\begin{table*}[h!]
\centering
\resizebox{\textwidth}{!}{
\begin{tabular}{l|cccccc}
\toprule
 &  MAE & MAE \tiny{(MAP only)} & MAE \tiny{Static PL} & MAE \tiny{changing PL} & Trend Acc. & CRPS\\
\midrule
MLP &9.85 \plusminus{0.44} & 4.11 \plusminus{0.01} & 8.88 \plusminus{0.45} & 13.76 \plusminus{0.40} & 0.83 \plusminus{0.03} & 7.43 \plusminus{0.22} \\
Neural Process & 8.32 \plusminus{0.18} & 4.63 \plusminus{0.06} & 6.83 \plusminus{0.26} & 14.31 \plusminus{0.22} & 0.89 \plusminus{0.00} & 4.92 \plusminus{0.66} \\
CLMU & 7.61 \plusminus{0.12} & 4.31 \plusminus{0.04} & 7.00 \plusminus{0.11} & 10.06 \plusminus{0.17} & 0.89 \plusminus{0.00} & 5.48 \plusminus{0.09} \\
SSM & 8.12 \plusminus{0.46} & 4.12 \plusminus{0.11} & 7.49 \plusminus{0.55} & 10.65 \plusminus{0.11} & 0.88 \plusminus{0.00} & 4.43 \plusminus{0.29} \\
TDT (ours) & \textbf{5.41 \plusminus{0.05} }& \textbf{3.88 \plusminus{0.12}} & \textbf{4.90 \plusminus{0.05}} & \textbf{7.47 \plusminus{0.08}} & 0.88 \plusminus{0.01} & \textbf{3.45 \plusminus{0.12}} \\

\bottomrule
\end{tabular}
}
\vspace{0.5em}
\caption{Digital twin model evaluation;  Transformer outperforms baselines in all metrics. } 
\label{tab:world_model_results}
\end{table*}

\subsection{Baselines}
We evaluate our approach against several established baselines for probabilistic dynamics modeling. Each baseline is configured with carefully tuned hyperparameters to ensure fair comparison:

\begin{itemize}
    \item \textbf{Multi-Layer Perceptrons (MLPs)} with Monte Carlo dropout approximate probabilistic forecasts by treating dropout as a Bayesian approximation technique, enabling uncertainty quantification through multiple forward passes during inference. The MLP baseline employs a three-layer architecture with hidden dimensions $[512, 256, 128]$, ReLU activation functions, and a dropout rate of $0.2$ applied after each hidden layer. The network flattens the input sequence and concatenates it with p-level control signals before processing through the fully connected layers.

    \item \textbf{Neural Processes} \cite{sun2024np} is a meta-learning approach that conditions on context observations to predict distributions over functions, enabling few-shot adaptation to new dynamical systems while maintaining uncertainty quantification. The implementation features a latent dimension of $128$, a hidden dimension of $256$, and employs separate encoder networks for context processing with three-layer architectures. The context encoder processes input features augmented with time indices, while the aggregator combines encoded representations across time steps. The decoder network generates both mean and variance predictions for each feature at each forecast timestep.

    \item \textbf{Conditional Legendre Memory Units (CLMUs)} \cite{li2022clmu,voelker2019lmu}  leverage orthogonal polynomial basis functions to capture long-term temporal dependencies through structured memory mechanisms. The CLMU baseline utilizes $2$ layers with memory dimension $64$, hidden dimension $128$, and incorporates p-level conditioning through a dedicated projection layer. Each LMU layer employs Legendre polynomial transition matrices with scaling parameter $\theta = 1.0$ and applies exponential smoothing with decay rate $0.9$ for stable memory updates. The output projection includes dropout with a rate $0.1$ for regularization.

    \item \textbf{State Space Models (SSMs)} \cite{ssm} represent dynamics through latent state evolution governed by linear or nonlinear transition functions, naturally incorporating temporal dependencies and enabling principled probabilistic inference over hidden states. The SSM baseline operates with state dimension $64$, hidden dimension $128$, and forecast horizon of $6$ steps. The state transition matrix is initialized as $0.9 \cdot I + 0.1 \cdot \mathcal{N}(0,1)$ to ensure stability, while the observation model employs a two-layer network with ReLU activation and dropout rate $0.1$. Stochastic sampling is achieved by injecting Gaussian noise with standard deviation $0.01$ during state transitions.

    \item \textbf{Transformers with sinusoidal positional embeddings  (TDT sin.)} \cite{vaswani2017attention} and \textbf{Transformers with rotary positional embeddings (TDT rot.)} \cite{su2024roformer}. TDT (rot.) leverages self-attention mechanisms enhanced with rotary position encoding (RoPE) that captures relative positional relationships through multiplicative rotations. The transformer model's attention mechanism allows the model to attend to relevant temporal patterns and input control, improving the time series prediction. Both transformer models have the architecture outlined in our methodology section.
\end{itemize}

All baselines are trained using the Adam optimizer with a learning rate $0.001$ and employ Monte Carlo sampling with $50$ forward passes for uncertainty quantification during inference.

\subsection{Metrics for evaluating digital twin}
\begin{itemize}
    \item  MAE All: Mean Absolute Error across all features
    \item MAE MAP: Mean Absolute Error for MAP (Mean Arterial Pressure) only
    \item MAE Static: MAE for samples with non-changing P-Levels over the course of 2 hours.
    \item MAE Dynamic: MAE for samples with dynamic p-levels
    \item Trend Acc: Trend direction accuracy for MAP. Trend is classified as (1) increasing if the slope of MAP over the predicted horizon is $\geq 2$, (2) decreasing if the slope $\leq 2$, and (3) flat otherwise.
    \item CRPS: Continuous Ranked Probability Score is a proper scoring rule \cite{gneiting2007strictly} for uncertainty quantification, calculated from 50 samples from the probabilistic predictions (samples $x, x'$) and the ground truth $y$ as in equation \ref{eq:crps}.
    
\begin{equation}
\label{eq:crps}
\text{CRPS}(\gF, y) = 
\int \left( \gF(x) - \mathbf{1}\{x \geq y\} \right)^2 dx = \mathbb{E}_x[|x - y|] - \frac{1}{2} \mathbb{E}_{x,x'}[|x - x'|] \quad 
\end{equation}
\end{itemize}

\subsection{Additional Visualizations.} 
\begin{figure}[h!]
    \centering
    \includegraphics[width=\linewidth]{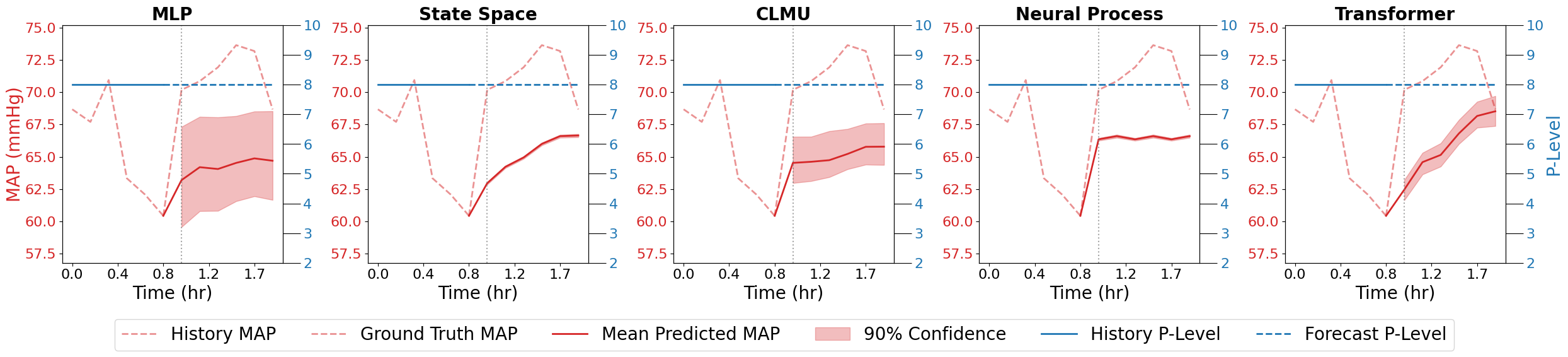}
        \includegraphics[width=\linewidth]{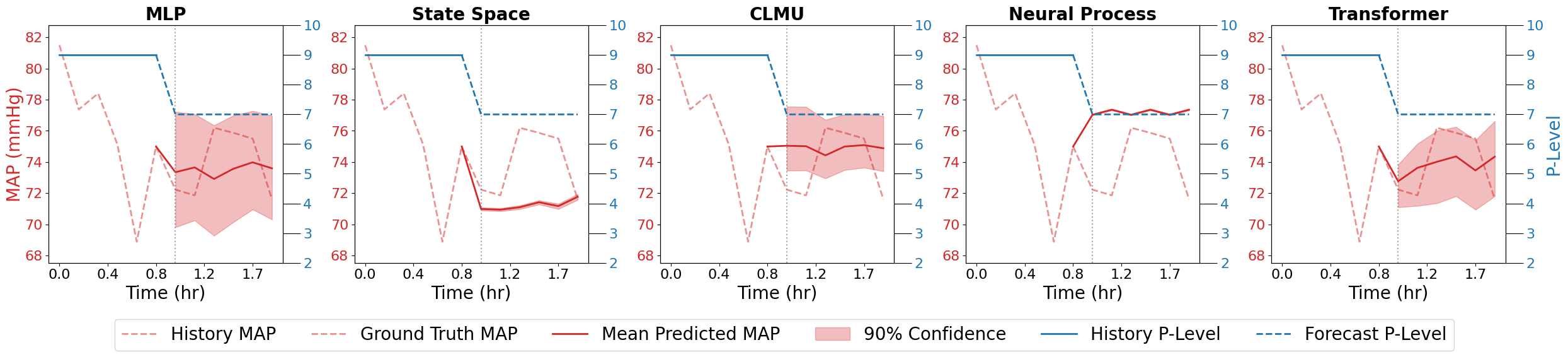}
       \includegraphics[width=\linewidth]{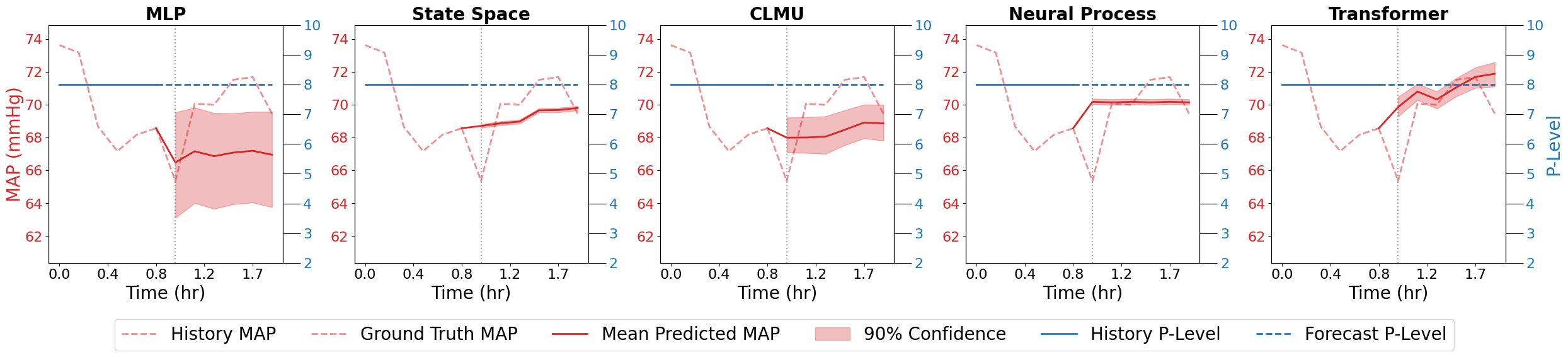}
    \vspace{-1em}
    \caption{Digital twin prediction visualization compared with baselines. The Transformer model is more accurate in reflecting response to P-level change and more expressive when capturing large changes in patient state, resulting in its higher accuracy.}
    \label{fig:dt_appendix}
\end{figure}

Please see figure \ref{fig:dt_appendix} for further qualitative examples of digital twin models.

\section{RL Experiment Settings and
Additional Results}
\label{sec:additional_rl}
\subsection{Implementation details and hyperparameters of CORMPO}
\label{sec:imp_det}
We first start our implementation by generating the replay buffer rewards shaped with $\lambda_1$ (ACP weight), and $\lambda_2$ (WS weight) parameters. ACP and WS takes the current state, $s_t$, as the stability condition and evaluates the change in action from current action, $a_t$, to the next action, $a_{t+1}$.
In parallel, we apply KDE on a randomly shuffled training split comprising 80\% of the full dataset and evaluate thresholds on a held-out 10\% validation set. Only for real dataset, we choose the threshold percentile as 35\% due to the distribution discrepancy between train and validation. For synthetic dataset, the optimal threshold is selected by searching across percentiles to on the validation set by minimizing the log-likelihood variance on the ID-labeled region resulting in 20\% for both.
For each query, we retrieve nearest neighbors from the training data and fit a kernel to compute log-density scores. 
Merging the two modules, we deploy KDE model and the selected threshold as the penalization on the rewards predicted by the dynamics model during \textsf{MBPO} training with the reward-shaped replay buffer.

%\subsection{Hyperparameters of \textsf{CORMPO}}

\paragraph{KDE Hyperparameters.}
We selected the RBF kernel for this method. The bandwidth and number of neighbors are selected based on the modeled distribution on the validation set. As we increase the bandwidth, the distribution starts to be sharper. So, we chose 1 as the bandwidth for KDE. We select 100 neighbors, as our training set includes 12051 samples.

\paragraph{CORMPO Hyperparameters.}
Our implementation of model-based policy optimization is largely borrowed by \cite{yu2020mopo} (see Table \ref{tab:mopo_param}) with the change of reward penalization in the dynamics model roll-outs. We select the hyperparameters of $\lambda_1$, $\lambda_2$, and $\lambda$ by evaluating each model in the TDT environment. Since there is no golden aggregation of the medical metrics, we pick the model with the best reward, with ACP and WS outperforming the baselines.
\begin{itemize}
    \item \textbf{Noiseless Dataset Experiment}: $\lambda_1=0.5$, $\lambda_2=0.3$, $\lambda=0.005$, $\text{epoch}=90$.
    \item \textbf{Noisy Dataset Experiment}: $\lambda_1=0.0$, $\lambda_2=0.0$, $\lambda=0.08$, $\text{epoch}=100$.
    \item \textbf{Real-life Dataset Experiment}:$\lambda_1=1.0$, $\lambda_2=0.0$, $\lambda=0.005$, $\text{epoch}=100$.
\end{itemize}

\begin{table*}[!t]
\centering
\begin{tabular}{p{0.45\linewidth} c}
\toprule
\textbf{Parameters} & \textbf{Value} \\
\midrule
Actor learning rate & $3\times10^{-4}$ \\
Critic learning rate & $3\times10^{-4}$ \\
Discount factor ($\gamma$) & $0.99$ \\
Target network update coefficient ($\tau$) & $0.005$ \\
Target entropy (often $-\text{action dimension}$) & $-1$ \\
Temperature optimizer learning rate & $3\times10^{-4}$ \\
Dynamics model learning rate & $1 \times10^{-3}$ \\
Dynamics ensemble size & $7$ \\
Holdout ratio & $0.2$ \\
Training epochs & $100$ \\
Steps per epoch & $1000$ \\
Evaluation episodes & $1000$ \\
Mini-batch size & $256$ \\
Model rollout horizon & $5$ \\
Rollout batch size & $10000$ \\
Rollout frequency & $1000$ \\
Real-to-model data sampling ratio & $0.05$ \\
\bottomrule
\caption{Base hyperparameters of our \textsf{CORMPO} implementation.}
\label{tab:mopo_param}
\end{tabular}
\end{table*}

\newpage

\subsection{Additional Results}
\label{sec:additional_results}

\begin{table*}[!ht]
\centering
\resizebox{\textwidth}{!}{%
\begin{tabular}{l|ccccc|cc}
\toprule
\textbf{Penalty type} & \textbf{Expert DT} & \textbf{BC} & \textbf{MBPO} & \textbf{MOPO} & \textbf{SVR} &
\multicolumn{1}{c}{\textbf{\textsf{CORMPO} \textit{KDE}}} &
\multicolumn{1}{c}{\textbf{\textsf{CORMPO} \textit{RealNVP}}} \\
\midrule
\textbf{Reward} & 0.557 & $0.175 \pm 0.118$ & $0.420 \pm 0.139$ & $0.373 \pm 0.129$ & $0.530 \pm 0.152$ & $\mathbf{0.687 \pm 0.106}$ & $0.516 \pm 0.126$ \\
\textbf{ACP}    & 1.79  & $0.068 \pm 0.012$ & $0.459 \pm 0.032$ & $0.984 \pm 0.020$ & $0.599 \pm 0.057$ & $\mathbf{0.018 \pm 0.007}$ & $0.258 \pm 0.036$ \\
\textbf{WS}     & 0.053 & $\mathbf{0.345 \pm 0.008}$ & $0.147 \pm 0.007$ & $0.042 \pm 0.006$ & $0.166 \pm 0.003$ & $0.173 \pm 0.007$ & $0.155 \pm 0.003$ \\
\bottomrule
\end{tabular}%
}
\caption{Comparison of \textsf{CORMPO} based on KDE, and \textsf{CORMPO} based on RealNVP against baseline models with 1000 episodes averaged over 5 seeds. Evaluation is completed in the noiseless environment setting. \textsf{CORMPO} with RealNVP does not outperform the baselines, showing moderate performance in all metrics. This result suggests further investigations on calibrating \textsf{CORMPO} to different density estimators.}
\label{tab:penalty_comparison}
\end{table*}

\begin{figure*}[tbh]
\floatconts
  {fig:parameter_sensitivity}%
  {\caption{Sensitivity analysis of \textsf{CORMPO} under (a) density thresholds,
  (b–c) penalty coefficients, and (d) regularization.}}%
  {%
    \centering

    % ---------- Row 1 ----------
    \begin{minipage}[t]{0.48\textwidth}
      \centering
      \includegraphics[width=\linewidth]{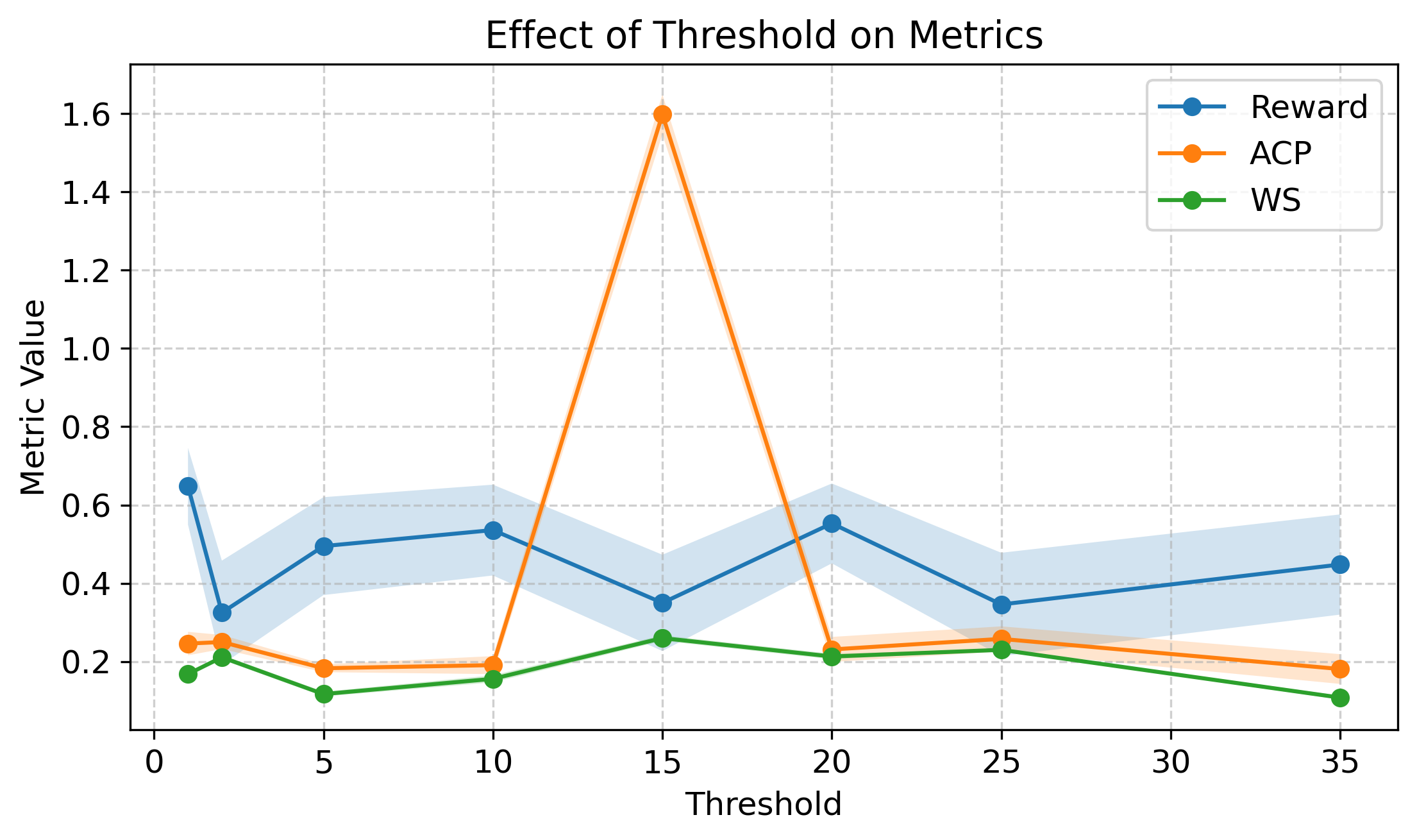}
      \vspace{4pt}
      {\small (a) Effect of threshold $\tau$. The x-axis is the \% of validation data flagged anomalous.}
      \label{fig:thr}
    \end{minipage}\hfill
    \begin{minipage}[t]{0.48\textwidth}
      \centering
      \includegraphics[width=\linewidth]{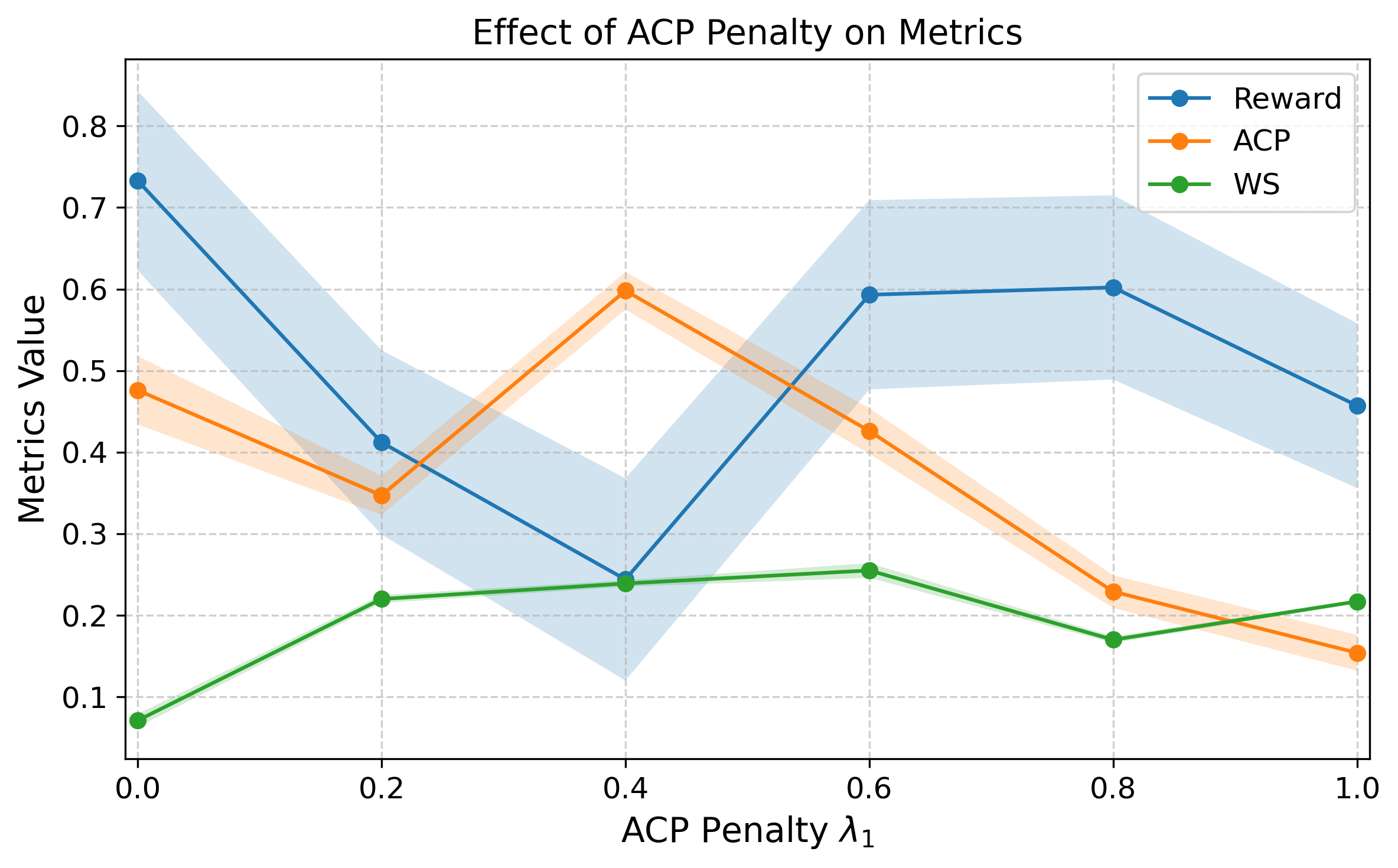}
      \vspace{4pt}
      {\small (b) Effect of ACP penalty $\lambda_{1}$ ($\lambda_{2}=0$, $\tau=0.005$).}
      \label{fig:acp}
    \end{minipage}

    \medskip

    % ---------- Row 2 ----------
    \begin{minipage}[t]{0.48\textwidth}
      \centering
      \includegraphics[width=\linewidth]{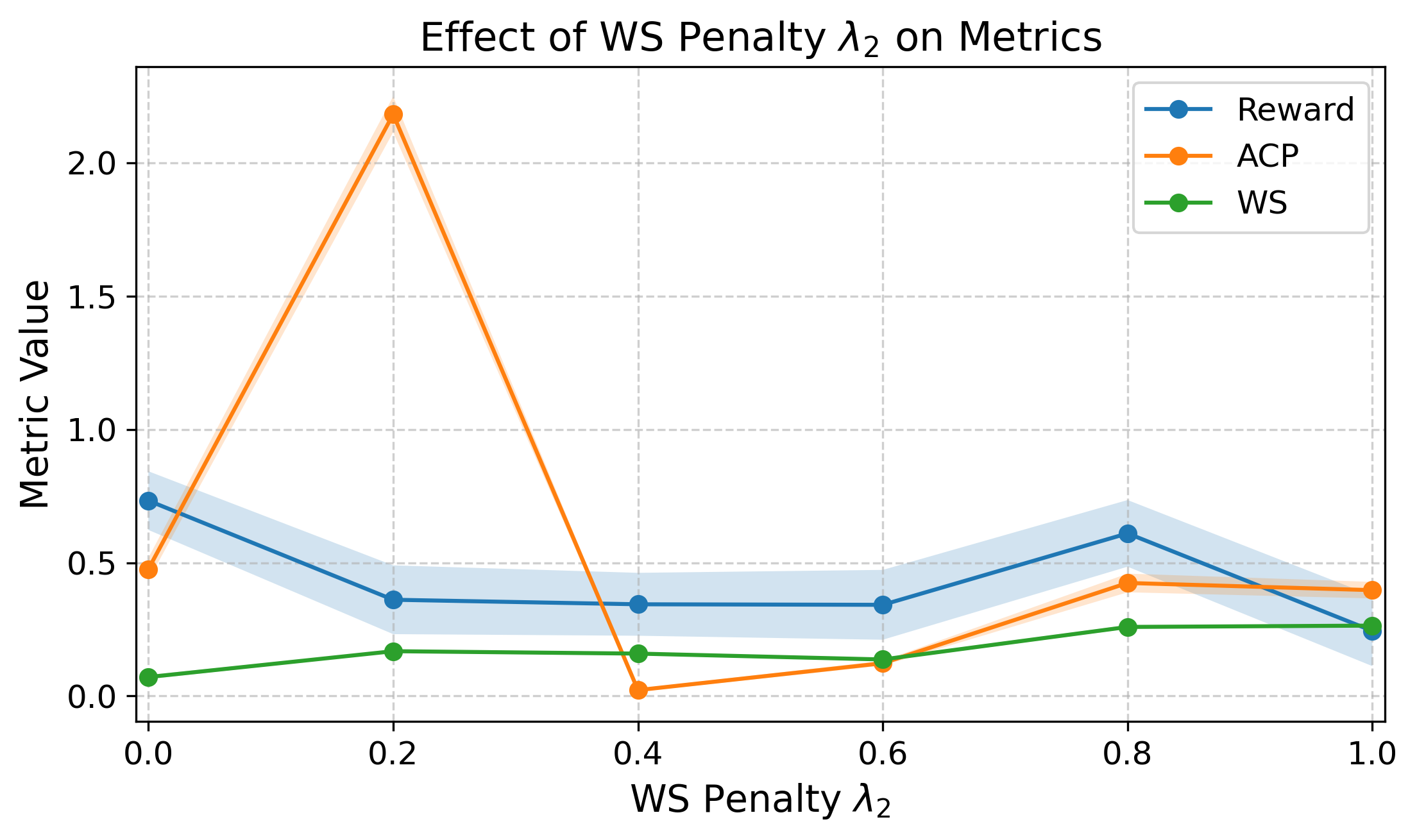}
      \vspace{4pt}
      {\small (c) Effect of WS penalty $\lambda_{2}$ ($\lambda_{1}=0$, $\tau=0.005$).}
      \label{fig:ws}
    \end{minipage}\hfill
    \begin{minipage}[t]{0.48\textwidth}
      \centering
      \includegraphics[width=\linewidth]{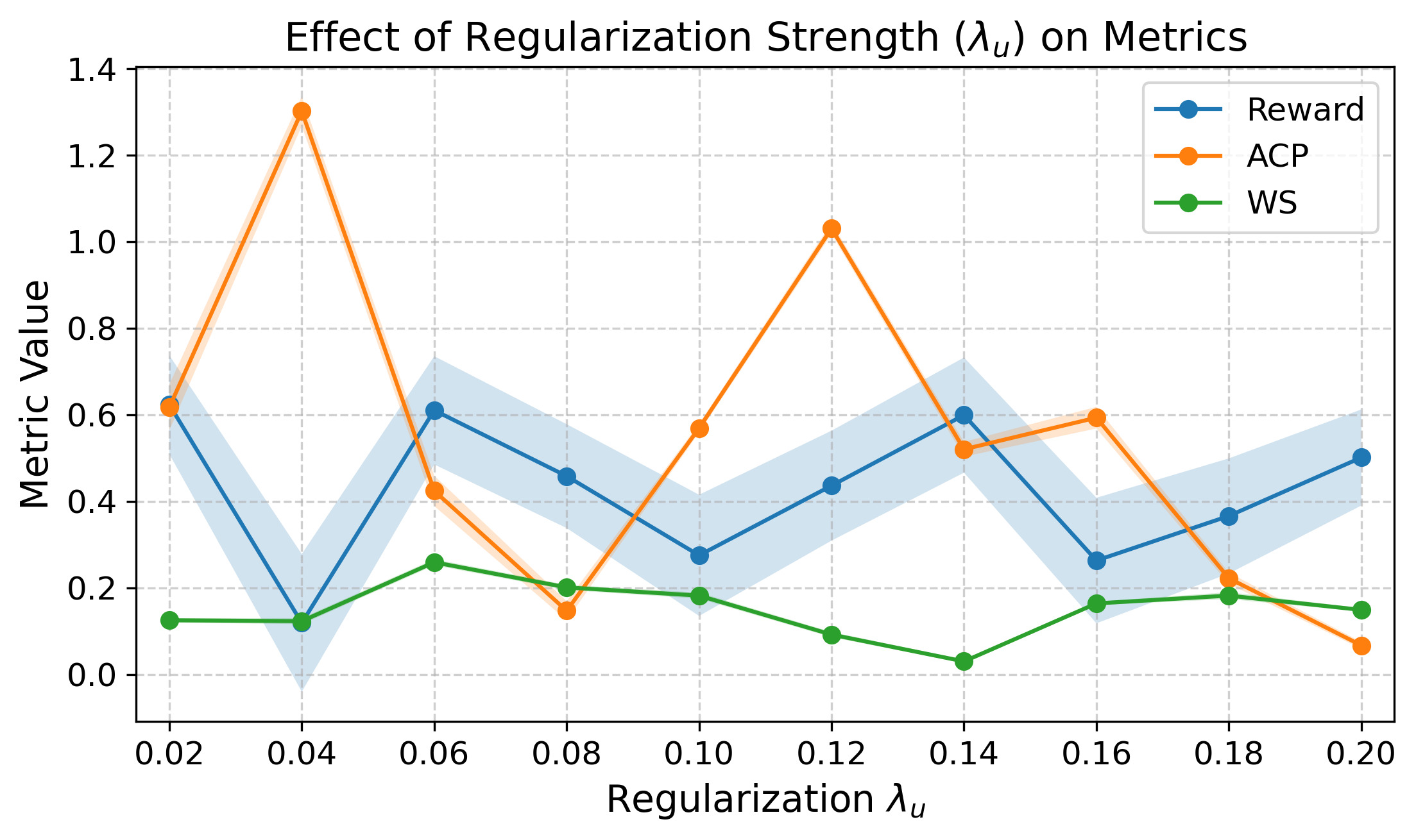}
      \vspace{4pt}
      {\small (d) Effect of regularizer $\lambda_{u}$ ($\lambda_{1}=0$, $\lambda_{2}=0.8$).}
      \label{fig:lambda_u}
    \end{minipage}
  }
\end{figure*}

%\clearpage
%\section{Qualitative Results}
\begin{figure*}[tbh]
    \centering
    % --- First plot ---
        \centering
        \includegraphics[width=\linewidth]{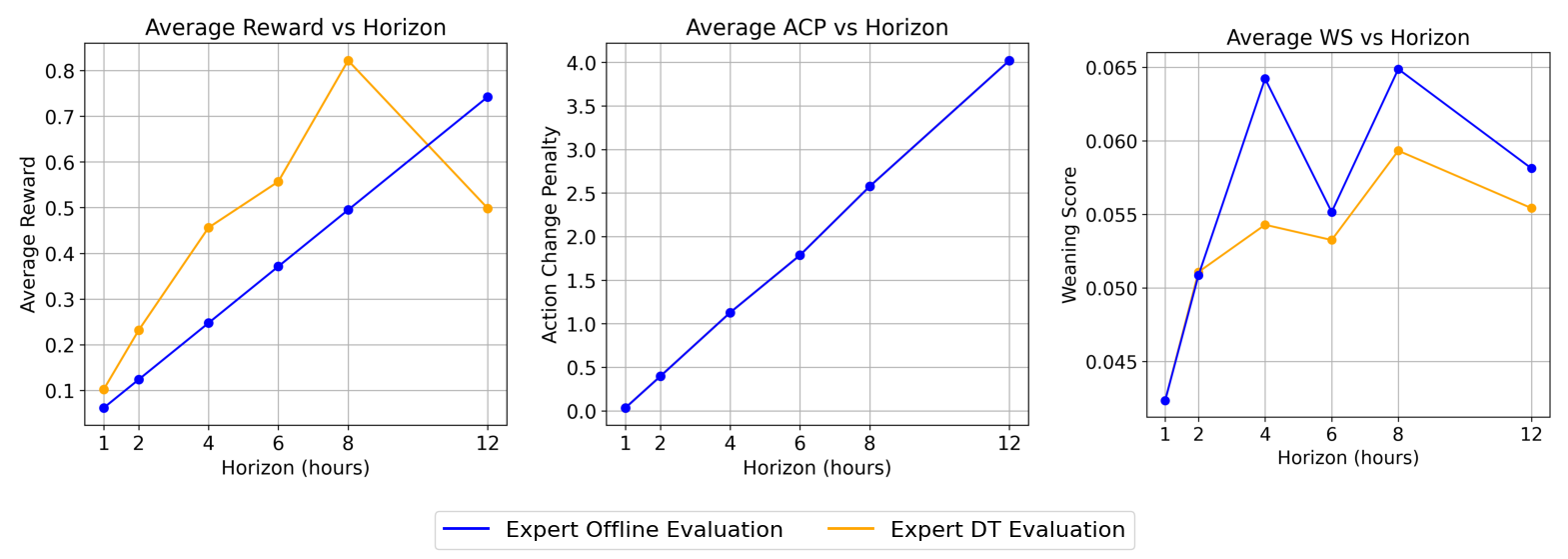}
        %\caption{Reward vs Threshold}
        \label{fig:reward_thr}
    \hfill
    \caption{Comparison of change in physiological reward, ACP, and WS metrics with respect to various evaluation horizons in hours. This experiment is done on the expert (clinician) decisions evaluated by our digital twin (orange) and directly on the offline patient outcomes (blue). 
    %For fairness to the compared offline RL policies, the reported expert evaluation on Tables 1-4 is with the digital twin given the real expert p-levels.
    We observe that our digital twin performs similar to the offline dataset. Noting that digital twin results in mostly higher physiological rewards, the gap between offline and digital twin evaluation increases within a tolerable bound, where the largest increase is 60\%. ACP remains the same since the compared actions are the real p-levels while only the next state differs. In WS, the digital twin depicts a small decrease, which is always bounded until the 12-hour horizon.
    }
    \label{fig:threshold_metrics}
\end{figure*}

\begin{figure*}[tbh]
    \centering
    \includegraphics[width=\linewidth]{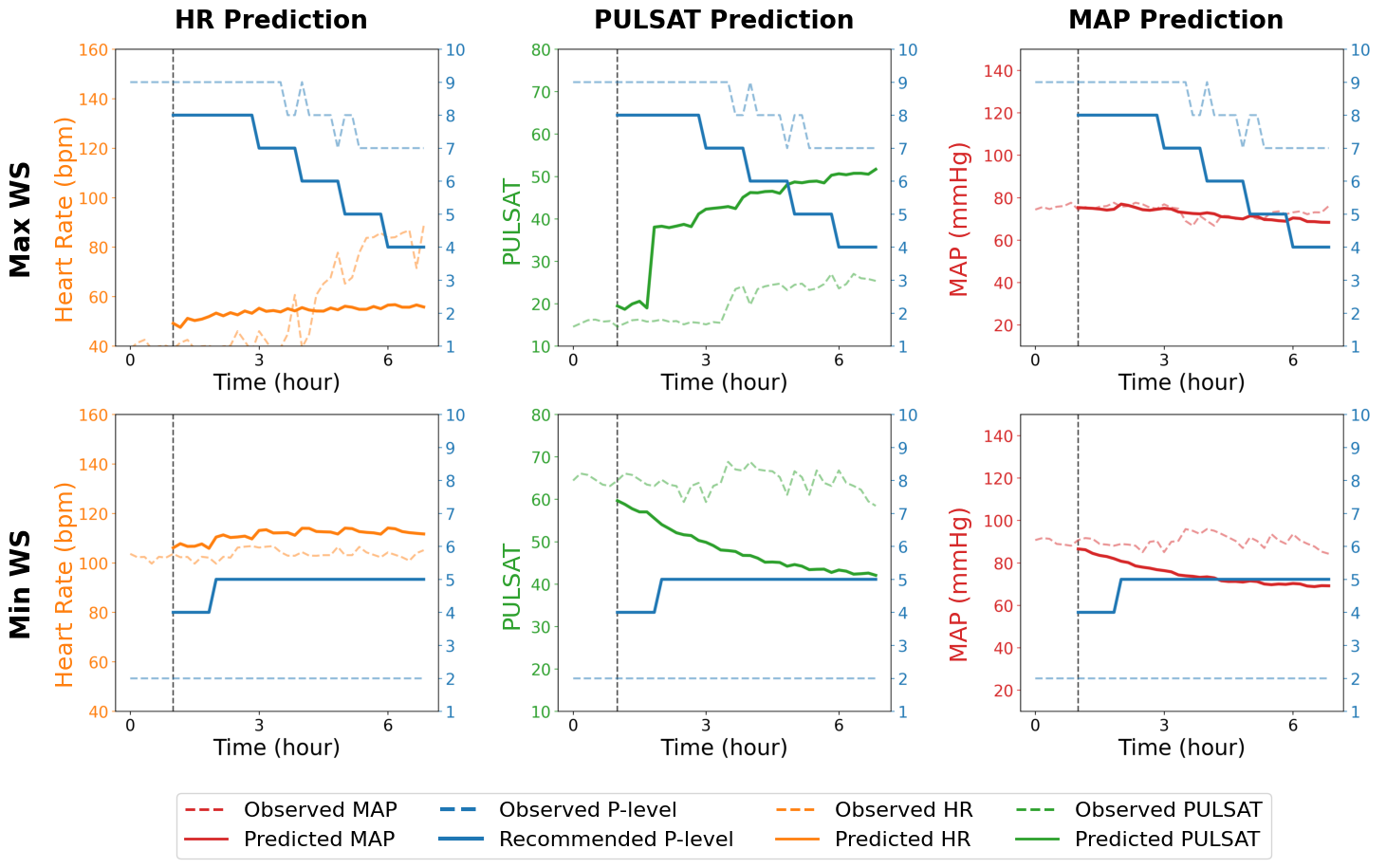}
    \caption{Trained on the real-life dataset, we compare our \textsf{CORMPO} against baselines in 6-hour digital twin rollouts. The upper row depicts high (0.75) and the lower row depicts low (-0.334) WS roll-outs with p-level recommendations on the real data. In the highest WS row, we predict all vitals to remain within safe hemodynamic regions (see Table \ref{tab:reward}), with HR and MAP showing stationary behavior and pulsatility rising above the critical threshold. We observe decaying pulsatility and MAP in the low WS case.}
    \label{fig:placeholder}
\end{figure*}

\begin{figure*}[tbh]
    \centering
    \includegraphics[width=\linewidth]{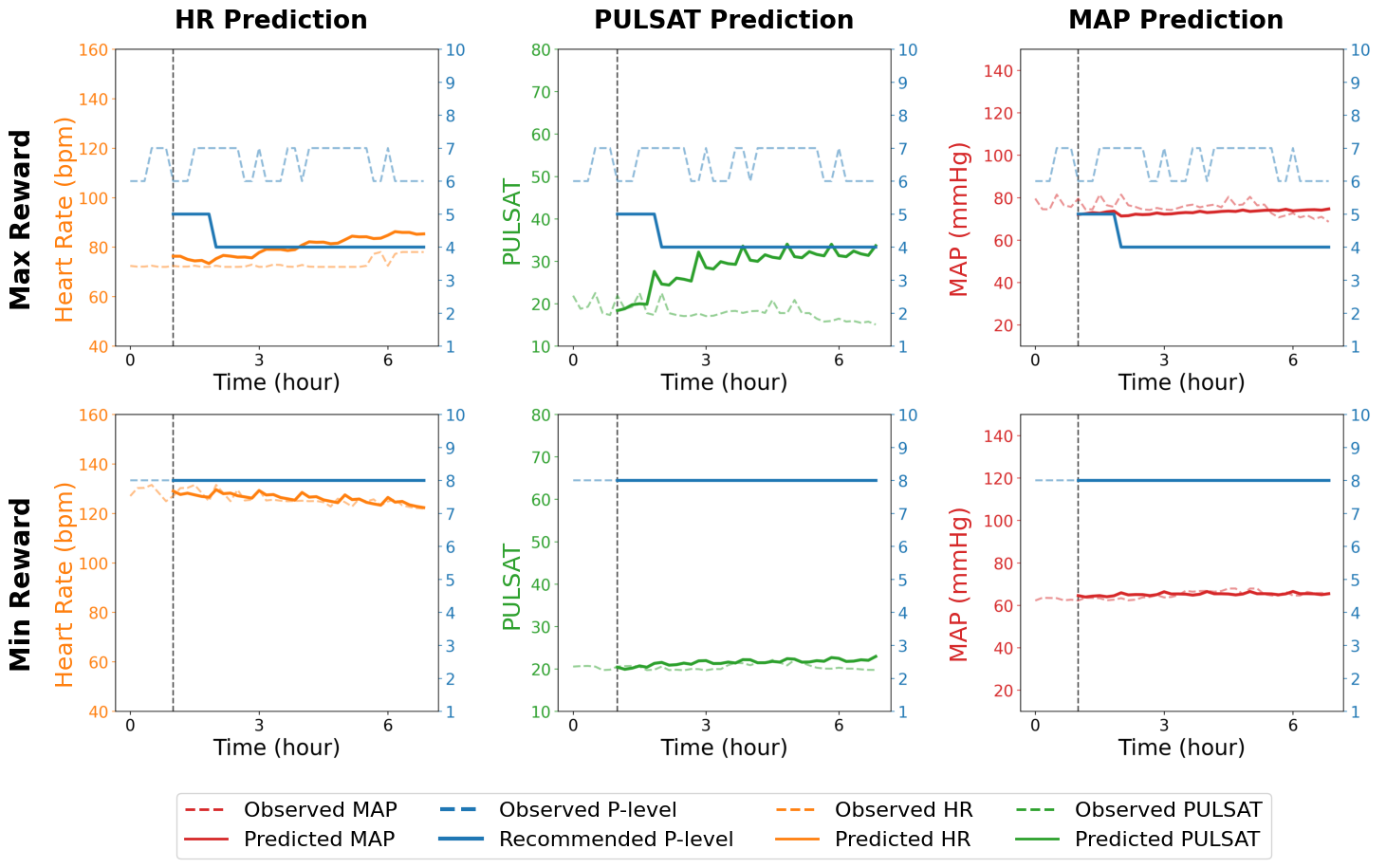}
    \caption{Trained on the real-life dataset, we compare our \textsf{CORMPO} against baselines in 6-hour digital twin rollouts. The upper row corresponds to the high physiological reward roll-outs (3.84), and the lower row to the low physiological reward roll-outs (–12.0). In the lower row, pulsatility and MAP are on the critical threshold. In the upper row, heart rate and pulsatility increase and stabilize away from the critical thresholds. As MAP is also predicted to remain stable, \textsf{CORMPO} initiates the weaning process.}
    \label{fig:placeholder}
\end{figure*}

\begin{figure}
    \centering
    \includegraphics[width=1\linewidth]{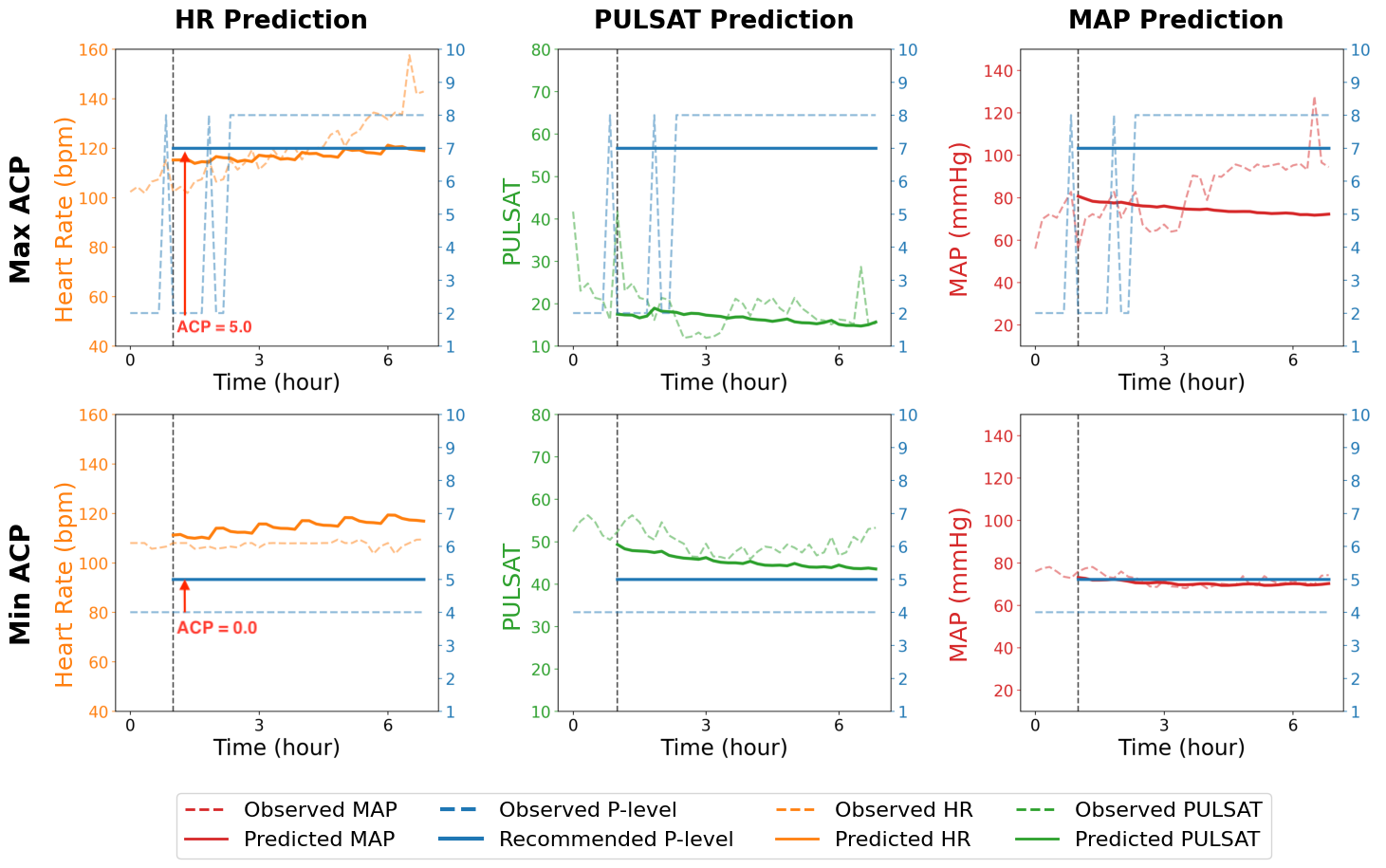}
    \caption{Trained on the real-life dataset, we compare our \textsf{CORMPO} against baselines in 6-hour digital twin rollouts. We depict high (5.00) and low (0.00) ACP roll-outs with p-level recommendations. To note, our ACP definition states $\text{ACP}_t = |a_{t+1} - a_t|$ only if $\text{ACP}_t > 2$ where $a_0$ is the action observed at time $t=1$ in the plots. Naturally, the upper row results in 5.0 $(|a_1-a_0| = 5)$ while the last row results in 0 ACP as indicated with the red arrow. In the lower row, with the p-level increased by 1 relative to the ground truth, the vitals remain stationary and away from the critical thresholds. In the upper row, ACP is 5.0 since the policy increases the p-level by 5 relative to the ground truth p-level. In this case, pulsatility and MAP decay down to the critical threshold.}
    \label{fig:placeholder}
\end{figure}

%%%%%%%%%%%%%%%%%%%%%%%%%%%%%%%%%%%%%%%%%%%%%%%%%%%%%%%%%%%%

%\newpage
%\input{secs/checklist}

\end{document}